\documentclass[10pt,twocolumn,letterpaper]{article}

\usepackage[pagenumbers]{iccv}      % To produce the REVIEW version
\usepackage{microtype}
\usepackage{graphicx}
\usepackage{booktabs} % for professional tables

\usepackage{amsmath}
\usepackage{amssymb}
\usepackage{mathtools}
\usepackage{amsthm}

\usepackage{url}
\usepackage{xcolor}         % colors
\usepackage{graphicx}
\usepackage{subcaption}   % For creating subfigures
\usepackage{adjustbox}
\usepackage{fp} % For floating-point calculations
\usepackage{xstring} % For string manipulation
\usepackage{wrapfig}
\usepackage{tabularx}
\usepackage{enumitem}
\usepackage{multicol}
\usepackage[most]{tcolorbox}
\usepackage{array}  % For wrapping text in table cells
\usepackage{multirow}

\usepackage{fvextra} % Enhanced verbatim functionalities
\tcbuselibrary{breakable}
\usepackage{soul}
\usepackage{tabularx}
\usepackage{multirow}

\newcommand{\laiontwob}{LAION-2B}
\newcommand{\laionfiveb}{LAION-5B}

\newcommand{\gptfouro}{GPT-4o}

\newcommand{\grade}{GRADE}

\newcommand{\sdclassicfirstckpt}{SD-1.1}
\newcommand{\sdclassic}{SD-1.4}
\newcommand{\sdtwo}{SD-2.1}
\newcommand{\sdvit}{SD-3 (2B)}
\newcommand{\sdxl}{SDXL}
\newcommand{\sdxlturbo}{SDXL-Turbo}
\newcommand{\sdxllcm}{SDXL-LCM}
\newcommand{\deepfloydM}{DeepFloyd-M}
\newcommand{\deepfloydL}{DeepFloyd-L}
\newcommand{\deepfloydXL}{DeepFloyd-XL}
\newcommand{\fluxschnell}{FLUX.1-schnell}
\newcommand{\fluxdev}{FLUX.1-dev}

\newcommand{\wimbd}{WIMBD}
\newcommand{\fid}{FID}
\newcommand{\recall}{Recall}
\newcommand{\precision}{Precision}
\newcommand{\inception}{Inception v3}
\newcommand{\clip}{CLIP}
\newcommand{\tvdgrade}{TVD\textsubscript{G}}

\newcommand{\numModelsTested}{12}
\newcommand{\numConceptsApproximated}{100}
\newcommand{\numPromptsPerSetting}{three}

\newcommand{\numPromptsInTotal}{600}
\newcommand{\numImagesPerPrompt}{100}

\newcommand{\AvgNumQuestionsPerConcept}{four}

\newcommand{\defbehthreshold}{80\%}

\newcommand{\numPromptDistributions}{2,430}
\newcommand{\numConceptDistributions}{405}

\definecolor{customGreen}{RGB}{0,128,0}
\definecolor{customBlue}{RGB}{0,0,255}
\definecolor{customRed}{RGB}{255,0,0}

\newcommand{\totalImagesPerModel}{60,000}

\usepackage[textsize=tiny]{todonotes}

\tcbset{
    myverbatimbox/.style={
        breakable,
        colback=blue!5!white,      % Background color
        colframe=blue!75!black,    % Frame color
        left=5pt,                  % Left padding
        right=5pt,                 % Right padding
        top=5pt,                   % Top padding
        bottom=5pt,                % Bottom padding
        enhanced,                  % Enable advanced features
        boxrule=1pt,             % Frame thickness
        fontupper=\small\ttfamily, % Font size and type (monospaced)
    }
}

\definecolor{iccvblue}{rgb}{0.21,0.49,0.74}
\usepackage[pagebackref,breaklinks,colorlinks,allcolors=iccvblue]{hyperref}
\usepackage[capitalize,noabbrev]{cleveref}

\def\paperID{4556} % *** Enter the Paper ID here
\def\confName{ICCV}
\def\confYear{2025}

\title{\grade{}: Quantifying Sample Diversity in Text-to-Image Models}

\author{
    Royi Rassin\\
    Bar-Ilan University\\
    \and
    Aviv Slobodkin\\
    Bar-Ilan University\\
    \and
    Shauli Ravfogel\\
    Bar-Ilan University\\
    ETH Zürich\\
    \and
    Yanai Elazar\\
    Allen Institute for AI\\
    University of Washington\\
    \and
    Yoav Goldberg\\
    Bar-Ilan University\\
    Allen Institute for AI\\
}

\begin{document}
\maketitle
\begin{abstract}

%%% original grade abstract
% Text-to-image (T2I) models are remarkable in generating realistic images based on textual descriptions. Yet, descriptions are inherently \emph{underspecified}: they do not include all possible attributes of the required image. This raises two key questions: Do models generate diverse outputs from underspecified prompts? How can we automatically measure diversity? To address these questions, we propose \grade{}: \textbf{Gr}anular \textbf{A}ttribute \textbf{D}iversity \textbf{E}valuation, an automatic method for quantifying sample diversity. \grade{} leverages the world knowledge embedded in large language models and visual question-answering systems to identify relevant concept-specific axes of diversity (e.g., ``shape'' and ``color'' for the concept ``cookie''). It then estimates frequency distributions of concepts and their attributes and quantifies diversity using entropy. We use \grade{} to measure the diversity of 12 models using 405 concept-attribute pairs, revealing that all models display limited variation. Further, we find that these models often exhibit \emph{default behaviors}, a phenomenon where a model consistently generates concepts with the same attributes (e.g., 98\% of the cookies are round). Lastly, we show that a key reason for low diversity is underspecified captions in training data. Our work proposes a modern, semantically-driven approach to measure sample diversity and highlights the stunning homogeneity in T2I models.

We introduce \grade{}, an automatic method for quantifying sample diversity in text-to-image models. Our method leverages the world knowledge embedded in large language models and visual question-answering systems to identify relevant concept-specific axes of diversity (e.g., ``shape'' for the concept ``cookie''). It then estimates frequency distributions of concepts and their attributes and quantifies diversity using entropy. We use \grade{} to measure the diversity of 12 models over a total of 720K images, revealing that all models display limited variation, with clear deterioration in stronger models. Further, we find that models often exhibit \emph{default behaviors}, a phenomenon where a model consistently generates concepts with the same attributes (e.g., 98\% of the cookies are round). Lastly, we show that a key reason for low diversity is underspecified captions in training data. Our work proposes an automatic, semantically-driven approach to measure sample diversity and highlights the stunning homogeneity in text-to-image models.\footnote{Project page and code: \texttt{\url{https://royira.github.io/GRADE}}.}

\end{abstract}    
\section{Introduction}
% \royi{need to add the big comparison figure again, with the diversity panel.}
% \begin{figure}[h]
%     \centering
%     {\includegraphics[width=1\columnwidth]{Figures/princess_20.pdf}}
%     \caption{\fluxdev{} generations for ``a princess at a children's party''. Despite different seed initializations, \royi{put here the deterioration figure and somewhere the rug images with diff prompts} princesses consistently appear white, with a pink dress and a tiara, though the prompt only mentions ``princess''.}
%     \label{fig:main_fig}
% \end{figure}

Text-to-image (T2I) models have the remarkable ability to generate realistic images based on textual descriptions. However, prompts are inherently \emph{underspecified} \citep{underspecified_t2i, dalle_seeing_double}, meaning they do not fully describe all attributes that appear in the resulting image. Often, we expect T2I models to \emph{produce diverse outputs} that represent the full spectrum of possible attributes. For example, when generating images of ``a cookie in a bakery'', we expect to see cookies with different shapes, colors, and textures, among other variations. But are current T2I models capable of generating diverse outputs? Evaluating diversity is inherently challenging because the set of possible attributes is virtually infinite. 
% Existing metrics, such as Fréchet Inception Distance (FID) \citep{fid_paper} and \precision{}-and-\recall{} \citep{precision_and_recall,improved_precision_and_recall} are supposed to measure diversity, but they are limited in their ability to capture granular forms of diversity, instead, they capture feature-level similarities. 
% These metrics also rely on a set of reference images, which is assumed to represent a diverse set of images; however, such set is often hard to obtain, and does not specify attributes of interest. 
% Our desiderata from a diversity metric is to be reference-free and human-interpretable.
Existing metrics, such as Fréchet Inception Distance (FID) \citep{fid_paper} and \precision{}-and-\recall{} \citep{precision_and_recall,improved_precision_and_recall} are supposed to measure diversity, but they are limited in their ability to capture granular forms of diversity, instead, they capture feature-level similarities. 
These metrics also rely on a set of reference images that typically reflects the training data distribution, which might not be diverse. Furthermore, such set is often hard to obtain, and does not specify attributes of interest. 
Our desiderata for a diversity metric is to be \textbf{reference-free, independent of a training data distribution, and human-interpretable.}

\begin{figure*}[t!]
    \centering
    \includegraphics[width=\textwidth]{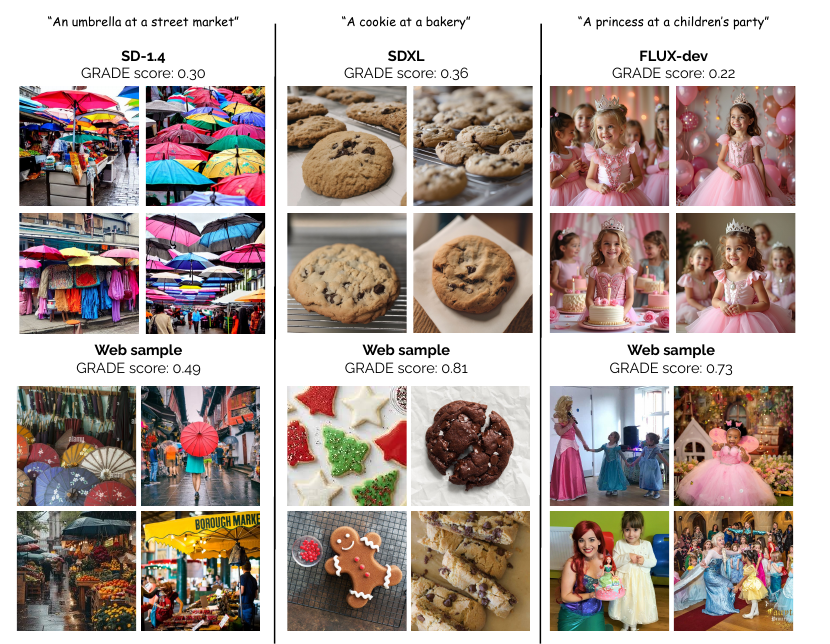}
    \caption{\textbf{\grade{} scores for T2I generations and corresponding web-search results, for three models and concepts.}}
    \label{fig:low_diversity_fig}
\end{figure*}
% \royi{need to add the big comparison figure again, with the diversity panel.}
% \begin{figure}[h]
%     \centering
%     {\includegraphics[width=1\columnwidth]{Figures/princess_20.pdf}}
%     \caption{\fluxdev{} generations for ``a princess at a children's party''. Despite different seed initializations, \royi{put here the deterioration figure and somewhere the rug images with diff prompts} princesses consistently appear white, with a pink dress and a tiara, though the prompt only mentions ``princess''.}
%     \label{fig:main_fig}
% \end{figure}

We propose \textbf{Gr}anular \textbf{A}ttribute \textbf{D}iversity \textbf{E}valuation (\grade{}), a method for measuring sample diversity in T2I models at a granular, \textbf{concept-dependent} manner, focusing on attributes, such as the \emph{shape} of a \emph{cookie} or the \emph{state} of an \emph{umbrella}. Our approach (illustrated in \cref{fig:pipeline_fig}) involves using a large language model (LLM) to generate prompts that elicit diverse outputs from T2I models. These prompts are accompanied by questions that tailor common, specific \emph{attributes}--relevant axes of diversity--for each concept (e.g., ``What is the shape of the cookie?'' and ``Is the umbrella open or close?''). We use a visual question-answering (VQA) model to extract attribute values from images using the questions. We then use an LLM to approximate the support of the concept and attribute, and map the VQA outputs to attribute values in the support. The result is a distribution over a concept and an attribute. We compute its normalized entropy and use it as our diversity score.

Using \grade{}, we determine that no model we test is particularly diverse, with the highest diversity score being 0.64 on a scale from zero to one. For example, \fluxdev{} (see \cref{fig:low_diversity_fig}) produces strikingly uniform images for ``a princess at a children's party'', scoring only 0.22; here, the princess is consistently white, with a dress and tiara--a phenomenon we call \emph{default behavior}. We explain such low scores from non-diverse images in the training data, often appearing with underspecified captions, which was previously explored in societal biases associations \cite{bias_amplification}.

Our contributions are threefold:
\begin{itemize}
    \item \textbf{A novel diversity evaluation method:} We introduce \grade{}, a fine-grained and interpretable method for evaluating diversity in T2I models that does not rely on reference images. We show \grade{} captures forms of diversity \fid{} and \recall{} do not.
    \item \textbf{Comparative diversity analysis:} Using \grade{}, we conduct an extensive study comparing the diversity of \numModelsTested{} T2I models, revealing that even the most diverse ones achieve low diversity and frequently exhibit \textit{default behaviors}. Our analysis uncovers negative correlation between model size and diversity.
    \item \textbf{Insights into influence of training data:} We demonstrate that underspecified captions in the training data contribute to low diversity of underspecified prompts. 
\end{itemize}

\section{Related Work} \label{sec:related_work}
Most diversity measurements are \emph{distribution-based}: a set of images generated by the evaluated model is compared to a reference set that captures the desired diversity, typically in feature-space, using a feature extractor such as \inception{} \citep{inceptionv3, inception_score} or \clip{} \citep{clipscore}.

Perhaps most popular, Fréchet Inception Distance (\fid{}) outputs a score representing both fidelity and diversity and is the standard for evaluating image generating models. However, it has multiple documented issues, like numerical sensitivity, data contamination, and biases \citep{fid_image_processing,fid_sample_size_bias, fid_biased_to_bias, fid_imagenet_bias, rethinking_fid}. \precision{}-and-\recall{} \citep{precision_and_recall} separated fidelity and diversity to two metrics. Additional metrics were proposed \citep{improved_precision_and_recall,density_and_coverage,attribute_based_diversity, faithful_synthetic_data}, which decouple between different properties and offer more interpretable methods.
Crucially, all these methods rely on a set of \textbf{diverse} reference images, by comparing the distribution of generated images to the reference set with the desired level of diversity. This can be the model's training data, or an established dataset, like ImageNet \citep{imagenet}.
However, acquiring reference images that faithfully reflect diversity is not straightforward and often requires using a feature extractor that was trained on similar data, to capture the similarities between the distributions. These requirements make it difficult to reproduce the results of previous work and maintain the integrity of the metrics as they are sensitive to data contamination, which could make them favor models that produce patterns similar to those seen in their training set, regardless of diversity \citep{fid_imagenet_bias}.

In addition to the significant requirement of obtaining a feature extractor and training data that match the target domain, previous metrics do not use fine-grained feature extractors, which can evaluate diversity over the semantics of images. Instead, they use ones that are trained over well-established datasets. As a result, they lack the ability to distinguish between two similar concepts that are different on a specific axis, like a color. For example, if we compare two nearly-identical images of a bottle, with only the color of the bottle as the difference, they would consider them very similar. However, our metric would capture such difference, as we show in \cref{app:comparing_grade_to_prev_metrics}.

Similar to \grade{}, Vendi Score \citep{vendiscore, cousins_of_vendiscore} is a reference-free metric, defined as the entropy of the eigenvalues of a user-provided similarity metric. However, it is sensitive to the choice of similarity function and is not fine-grained and interpretable in natural text.

OpenBias \citep{openbias} is an automatic bias detection framework that leverages LLMs and VQAs to identify an open set of biases in T2I models, focusing on assessing fairness by detecting novel bias patterns. Although \grade{} takes a similar approach to this method, its purpose is inherently different: to quantify and interpret the sample diversity of T2I models by measuring attribute variability and providing a reliable diversity score over the concepts tested.

\section{GRADE: Measuring Diversity in Models} \label{sec:pipeline}

\begin{figure*}[t!]
    \centering
    \includegraphics[width=\textwidth]{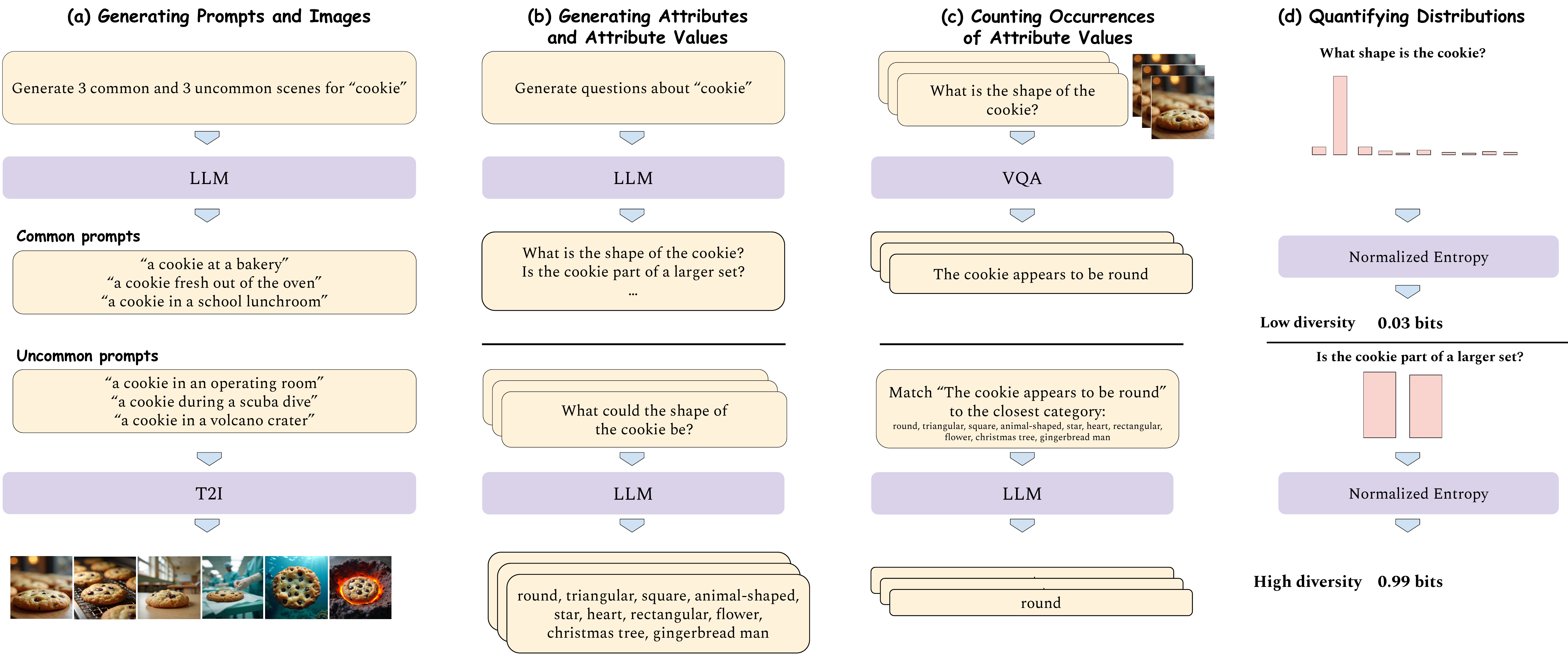}
    \caption{\textbf{Workflow of \grade{} using “cookie” as input.} (a) Generate prompts that mention “cookie” without specifying its attributes, and use them to generate images. (b) Formulate attribute-related questions and extract responses from the images using a VQA model. (c) Produce attribute values and map the responses to these values. (d) Quantify the diversity of the resulting attribute distributions.}
    \label{fig:pipeline_fig}
\end{figure*}

\subsection{Approach} \label{sec:problem_description}

We seek to quantify the variability of images produced by a T2I model for a given concept \(c\) when the prompt underspecifies certain attributes. Concretely, let \(C\) be a random variable representing possible \emph{concepts} (\emph{e.g.}, “cookie”) and let \(A\) be a random variable representing \emph{attributes} (\emph{e.g.}, “shape”). An attribute \(A = a\) may take values in a set \(\mathcal{V}_{c}^{a}\), denoting the ways in which \(a\) can manifest for concept \(c\) (\emph{e.g.}, a cookie’s shape could be “round” or “square”).

To characterize the probability of observing an attribute value \(v \in \mathcal{V}_{c}^{a}\) in a generated image, we define the \emph{concept distribution}:
\begin{equation}
    P_{V \mid a,c}(v) \;=\; P\bigl(V = v \,\mid\, A = a,\, C = c\bigr).
\end{equation}
Ideally, one would generate \emph{all} possible images of \(c\) to empirically determine the frequency of each value \(v\). However, both the conceptual and attribute spaces can be immense, making exhaustive enumeration infeasible. Moreover, identifying which attributes apply to a given concept necessitates world knowledge (for instance, “open or closed” is relevant for an umbrella but not for a cookie).

Instead, we approximate the attribute set \(\mathcal{V}_{c}^{a}\) with \(\tilde{\mathcal{V}}_{c}^{a}\) based on language-model-derived world knowledge. Furthermore, we define a set of \emph{underspecified prompts} \(\mathcal{P} = \{ p_1, p_2, \ldots, p_n\}\), each referencing the concept \(c\) while leaving the targeted attributes unspecified. We then obtain a \emph{multi-prompt distribution}:
\begin{equation}
    \tilde{P}_{V \mid a,c}(v)
    \;=\;
    \frac{1}{n}\,\sum_{i=1}^{n}\,
    P\!\Bigl(V = v \,\mid\, A = a,\, C = c,\, p_i\Bigr),
\end{equation}
which reflects, across multiple prompts, how frequently the T2I model generates the attribute value \(v\) for concept \(c\).

To measure diversity, we compute the normalized entropy of \(\tilde{P}_{V \mid a,c}\):
\begin{equation}
    \hat{H}\bigl(\tilde{P}_{V \mid a,c}\bigr)
    \;=\;
    \frac{H\bigl(\tilde{P}_{V \mid a,c}\bigr)}
         {\log_2 \bigl|\tilde{\mathcal{V}}_{c}^{a}\bigr|},
\end{equation}
where \(H(\cdot)\) is the Shannon entropy and \(\lvert \tilde{\mathcal{V}}_{c}^{a}\rvert\) is the cardinality of the approximate attribute-value set. By definition, \(\hat{H}\) ranges from \(0\) (all images collapse onto a single attribute value) to \(1\) (the attribute values are evenly distributed). We will refer to \(\hat{H}\) simply as the “entropy” for brevity.

Although we focus on \emph{multi-prompt distributions}, one can also examine \emph{single-prompt distributions}, which measure how a single prompt \(p \in \mathcal{P}\) distributes across the attribute values in \(\tilde{\mathcal{V}}_{c}^{a}\). Averaging these metrics across concepts and attributes yields a global measure of a T2I model’s diversity.

\subsection{Method}

Our proposed \grade{} pipeline comprises four steps (Fig.~\ref{fig:pipeline_fig}):

\textbf{(a) Generating images of a concept \(\mathbf{c}\).}
We first design two kinds of underspecified prompts for each concept \(c\):
\textit{common prompts}, which situate \(c\) in familiar or high-frequency scenarios that often appear in web-scale training corpora, and \textit{uncommon prompts}, which deliberately embed \(c\) in rare or surprising contexts. Common prompts (\emph{e.g.}, “a cookie during Christmas festivities”) may highlight typical attribute-value associations (such as tree-shaped cookies), whereas uncommon prompts (\emph{e.g.}, “a cookie in a volcano crater”) test whether the model defaults to certain “usual” attributes even under contextually unusual conditions. This dichotomy reveals whether certain attributes (like shape) are persistently tied to \(c\) despite substantial context shifts.

\textbf{(b) Generating attributes and their values.}
Next, we identify which attributes are relevant to each concept, such as “color,” “shape,” or “state” (open/closed). We query an LLM with the target concept \(c\) to produce a list of candidate attributes and corresponding questions (\emph{e.g.}, “What is the shape of the cookie?”). For each question, we derive a set of plausible answers (\emph{e.g.}, “round,” “square”), merged into \(\tilde{\mathcal{V}}_{c}^{a}\) by unifying semantically similar terms. This step ensures we capture domain-appropriate attributes for each concept and avoid missing frequently occurring variations. \cref{tab:sample_of_concepts_and_attributes} provides a representative sample of the generated concepts, attributes, and their corresponding candidate values, illustrating how the support \(\tilde{\mathcal{V}}_{c}^{a}\) is constructed.

\textbf{(c) Counting occurrences of attribute values in images.}
For each generated image, we obtain a \emph{VQA-based} natural-language answer to the attribute question. An LLM then maps this free-form answer to one of the entries in \(\tilde{\mathcal{V}}_{c}^{a}\). If no valid match is found—\emph{e.g.}, because the image fails to depict concept \(c\) or the answer goes beyond the specified set—the response defaults to “none of the above.” We discard such cases from the normalized frequency distribution and tally the rest of the attribute values across images. Repeating this for all prompts in \(\mathcal{P}\) yields our approximated multi-prompt distribution \(\tilde{P}_{V \mid a,c}(v)\).

\textbf{(d) Quantifying distributions.}
\grade{} then applies the normalized entropy measure to each concept-attribute distribution to quantify diversity. Higher entropy indicates that a T2I model spreads its generations more uniformly across \(\tilde{\mathcal{V}}_{c}^{a}\), whereas lower entropy suggests mode collapse toward a particular attribute value. We further aggregate all concept-attribute pairs into a model-level diversity score. 

\paragraph{Implementation details.}
In step (a), we generate \(\numPromptsPerSetting{}\) common and \(\numPromptsPerSetting{}\) uncommon prompts per concept, each yielding \(\numImagesPerPrompt{}\) images for a T2I model. In step (b), we typically have \(\AvgNumQuestionsPerConcept{}\) attributes per concept. Our pipeline uses \gptfouro{}\;\cite{gpt4} (\texttt{gpt-4o-2024-08-06}) with temperature \(0\) and a \(\max\) token limit of \(1{,}000\). Although described in multiple sub-steps, we leverage structured output techniques \citep{openai_structured_outputs_api} to streamline the question-answering and attribute-value mapping into a unified process. Full prompt details are provided in Appendix~\ref{app:prompts_in_grade}.

The cost of estimating a multi-prompt distribution is approximately \( \$ 0.75 \), and a single prompt distribution is \( \$0.12 \), using batch inference. In our experience, wait time is several minutes. Images were generated using an A100-80GB.

\begin{table}[t]
  \centering
  \small
  \begin{tabularx}{\columnwidth}{@{} >{\raggedright\arraybackslash}p{1cm} >{\raggedright\arraybackslash}p{2.5cm} >{\raggedright\arraybackslash}X @{}}
    \toprule
    \textbf{Concept} & \textbf{Attribute} & \textbf{Attribute Values} \\
    \midrule
    Teapot   & What shape is the teapot?                   & rectangular, spherical, oval, round, square, cylindrical \\
    Person   & Does the person appear to be alone or with others? & alone, with others \\
    Suitcase & Is this a vintage suitcase?                  & yes, no \\
    Bear     & What species of bear is depicted in the image?  & polar bear, black bear, sloth bear, grizzly bear, sun bear, panda bear \\
    \bottomrule
  \end{tabularx}
  \caption{\textbf{Sample of concepts, attributes, and attribute values.} Each concept-attribute pair is a multi-prompt distribution. A larger sample can be viewed in \cref{app:extended_data_overview}.}
  \label{tab:sample_of_concepts_and_attributes}
\end{table}

\section{Validating \grade{}} \label{subsec:validating_grade}

We extensively validate each component of \grade{}---except step (d), which only involves applying the normalized entropy formula and is conceptually straightforward. Our goal is to confirm that the generated prompts, attributes, and extracted attribute values are accurate, and that the answers provided by our underlying VQA model reliably match human judgment. Below, we detail our validation procedure, which involves both expert review and human annotation on 2{,}800 images. While these checks are performed here to establish confidence in \grade{}, they are not required every time the method is applied.

\textbf{(a) Prompt validity.} We first scrutinize all \numPromptsInTotal{} automatically generated prompts to ensure that each indeed mentions the intended underspecified concept (i.e., the prompt includes the concept but does not specify its attributes). To confirm the distinction between \emph{common} and \emph{uncommon} prompts, we extract all nouns in the prompt and measure their co-occurrence frequencies in \laionfiveb{}~\citep{laion-5b-paper}, using the large-scale dataset tool \wimbd{}~\citep{wimbd}. On average, the nouns in our common prompts appear 30,655 times in \laionfiveb{}, whereas those in our uncommon prompts appear only 956 times. These counts verify that the prompts are appropriately categorized. Note that we do not evaluate the quality of the generated images at this stage, as image fidelity depends on the T2I model rather than \grade{} itself.

\textbf{(b) Attribute and attribute-value validity.} Next, we validate that each of the \numConceptDistributions{} attribute-focused questions indeed corresponds to a visually discernible property of the concept (e.g., “What is the shape of the cake?”) and that no redundant or synonymous attribute values (e.g., “round” versus “circle”) coexist in the same support set. We then examine whether the support set adequately represents all attribute values extracted from the crowdsourced evaluation (see step (c) below). Specifically, we analyze every instance labeled “none of the above” (i.e., no valid match to our support). Among 1{,}000 sampled examples, 115 (11.5\%) fell into this category. Of these, only 3 (2.6\% of the 115) were true mismatches where the correct attribute was absent from our support. In 92 cases (80\%), the T2I model simply failed to follow the prompt by omitting the target concept. In the remaining 20 cases (17.3\%), either the VQA model or human annotators provided an incorrect answer. These low error rates confirm that our question sets and attribute-value supports are comprehensive.

\textbf{(c) Answerability of the questions.} Lastly, we verify that \gptfouro{} can correctly answer the attribute-based questions generated in step~(b). We conduct two Amazon Mechanical Turk (AMT) studies: (i) a broad assessment of 1{,}000 images (sampled from \numModelsTested{} T2I models), and (ii) a focused evaluation on a single multi-prompt distribution (“What is the shape of the cake?”) using 1{,}800 images from three representative models: \sdclassic{}~\citep{stable_diffusion}, \sdxlturbo{}~\citep{sdxl_turbo}, and \fluxdev{}~\citep{flux_announcement}. In both studies, we display to the workers (1) the image, (2) the question, and (3) the set of possible attribute values (including “none of the above”). Each example is labeled by three workers, and we take the majority vote. In the broad assessment, \gptfouro{}'s answers match the majority decision in \textbf{90.2\%} of the 1{,}000 examples. In the second experiment, overall agreement rises to \textbf{92.8\%} across all 1{,}800 “cake” images, with model-specific agreements of 88\% (\sdclassic{}), 91.2\% (\fluxdev{}), and 99.5\% (\sdxlturbo{}). These results establish that \gptfouro{} is a reliable VQA backbone for \grade{}. Additional details on our human evaluation setup can be found in \cref{app:human_eval}.

\subsection{Comparing \grade{} to Previous Metrics} \label{subsec:correlation_to_existing_metrics}

\begin{table}[t]
\centering
% \small
\begin{tabular}{@{}llrr@{}}
\toprule
\textbf{Model} & \textbf{Dataset} & \textbf{TVD-\fid{}} & \textbf{TVD-R} \\
\midrule
\sdclassicfirstckpt{} & \laiontwob{}  & \(0.12\) & \(-0.15\) \\
\sdclassic{}          & \laiontwob{}  & \(0\)    & \(-0.20\) \\
\sdtwo{}              & \laionfiveb{} & \(0\)    & \(-0.19\) \\
\bottomrule
\end{tabular}
\caption{\textbf{PCC between \grade{} and traditional metrics paired with \clip{}.} \fid{} has near-zero or low correlation with TVD, while \recall{} (R) exhibits a negative correlation. These results indicate that the attribute-focused distributions captured by \grade{} contrast sharply with what existing feature-level metrics measure.}
\label{tab:correlation_to_existing_metrics_clip}
\end{table}

After establishing the reliability of \grade{} in \cref{subsec:validating_grade}, we now compare it to two widely used metrics: \fid{}~\citep{inceptionv3} and \recall{}~\citep{precision_and_recall}, both of which assume feature-level distributions or predefined references. Specifically, we modify \grade{} to operate as a reference-based metric by replacing its entropy term with Total Variation Distance (TVD). This variant, compares an estimated distribution to a corresponding reference distribution from LAION~\citep{laion-5b-paper} in a manner analogous to how \fid{} and \recall{} rely on reference datasets.

\cref{tab:correlation_to_existing_metrics_clip} reports Pearson Correlation Coefficients (PCC) between TVD and the classical metrics. We observe that \fid{} is nearly uncorrelated with TVD, while \recall{} is negatively correlated. This divergence arises because \fid{} and \recall{} summarize distributions in feature space (e.g., via \clip{}~\citep{clipscore}), which may overlook fine-grained attribute variations (e.g., different shapes for a concept like ``cookie''). In contrast, \grade{} explicitly models attribute values grounded in human-understandable questions (e.g., ``Is the cookie round or square?''), thus capturing concept-level diversity that global feature statistics fail to discern.

These findings echo our human-based validations, which confirm that \grade{} effectively measures semantic-level variation missed by \fid{} or \recall{}. Further experiments, including analogous analyses using \inception{}~\citep{inceptionv3} as a feature extractor, reinforce these conclusions (see \cref{app:comparing_grade_to_prev_metrics}). Overall, the gap between traditional reference-based metrics and \grade{} illustrates the benefit of focusing on concept-specific attributes when assessing generative diversity.

\section{Comparing Diversity of Models} \label{sec:comparing_t2i_models}
We use \grade{} to estimate the diversity of popular T2I models.
We begin with an overview of our setup and then present the results.

\textbf{Data and distributions overview. } For each model, we estimate distributions over \numConceptsApproximated{} common concepts such as ``cookie'' and ``suitcase'' and attributes such as ``shape'' and ``color''.
Each concept is linked to \AvgNumQuestionsPerConcept{} questions on average. In total, there are \numConceptDistributions{} multi-prompt distributions and \numPromptDistributions{} single prompt distributions, consisting a total of \totalImagesPerModel{} images per model.

\textbf{T2I models.} We use \numModelsTested{} models from three families. 
\textbf{IF-DeepFloyd} \citep{DeepFloydIF} includes \deepfloydM{}, \deepfloydL{}, and \deepfloydXL{}. 
\textbf{Stable Diffusion} \citep{stable_diffusion, sdxl, sdxl_turbo, sdxl_lcm, stable_diffusion_vit} 
includes \sdclassicfirstckpt{}, \sdclassic{}, \sdtwo{}, \sdxl{}, \sdxlturbo{}, \sdxllcm{}, and \sdvit{}. 
Finally, \textbf{FLUX} \citep{blackforestlabs2024schnell, blackforestlabs2024dev} includes 
\fluxschnell{} and \fluxdev{}. All models were used with the default \texttt{Diffusers} library \citep{diffusers2023} settings.

% \textbf{T2I models. } We use \numModelsTested{} models from three different families shown in \cref{tab:t2i_model_families}. All models were used with their default hyperparameters as in the \texttt{Diffusers} library \citep{diffusers2023}.

% \begin{table}[h]
% \centering
% \scriptsize
% % \small
% \begin{tabularx}{\linewidth}{@{} p{2cm} >{\raggedright\arraybackslash}X @{}}
% \toprule
% \textbf{Family} & \textbf{Models} \\ 
% \midrule
% \textbf{IF-DeepFloyd} & 
% \deepfloydM{}, \deepfloydL{}, \deepfloydXL{} \citep{DeepFloydIF} \\ 
% \textbf{Stable Diffusion} & 
% \sdclassicfirstckpt{}, \sdclassic{}, \sdtwo{} \citep{stable_diffusion}, \sdxl{} \citep{sdxl}, \sdxlturbo{} \citep{sdxl_turbo}, \sdxllcm{} \citep{sdxl_lcm}, \sdvit{} \citep{stable_diffusion_vit} \\ 
% \textbf{FLUX} & 
% \fluxschnell{} \citep{blackforestlabs2024schnell}, \fluxdev{} \citep{blackforestlabs2024dev} \\ 
% \bottomrule
% \end{tabularx}
% \caption{\textbf{The \numModelsTested{} T2I Models grouped by family.}}
% \label{tab:t2i_model_families}
% \end{table}

\subsection{Results} \label{subsec:results}

\begin{figure*}
    \centering    

\includegraphics[width=\textwidth]{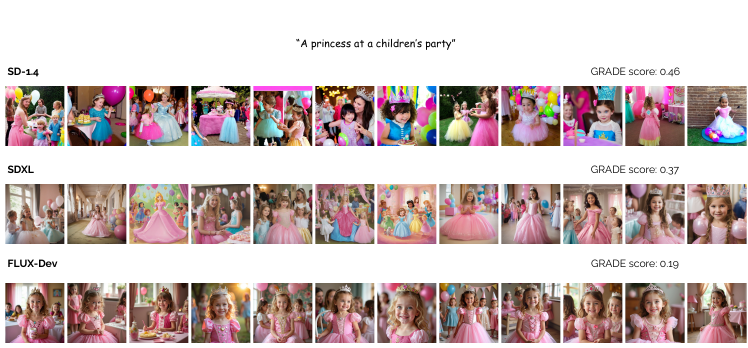}
    \caption{
    \textbf{Images generated with the prompt ``a princess at a children's party'' show differences in model diversity.} From top to bottom, \sdclassic{} (most diverse), \sdxl{}, and \fluxdev{} (least diverse). Although none are highly diverse, there is a marked difference between them. Specifically, diversity is reduced in attributes such as the ethnicities of depicted people, colors of dresses, and overall backgrounds.
    }
    \label{fig:princess_diversity_fig}
\end{figure*}

\begin{table}[h]

\centering
\resizebox{\columnwidth}{!}{
% \resizebox{\columnwidth}{!}{
\begin{tabular}{@{}lcc@{}}
\toprule
& \multicolumn{2}{c}{\textbf{GRADE Score $\uparrow$}} \\
\cmidrule(lr){2-3}
\textbf{Model} & \textbf{Multi-prompt} & \textbf{Single-prompt} \\
\midrule
\deepfloydM{} & \( \mathbf{0.64}\) &  \( 0.49\)\\
\deepfloydL{} & \( 0.62 \) & \(0.47\) \\
\deepfloydXL{} & \(0.61\) & \(0.46 \)\\
\sdclassicfirstckpt{} &  \( \mathbf{0.64}\) & \( \mathbf{0.54}\)\\
\sdclassic{} & \( \mathbf{0.64}\) & \( 0.53\) \\
\sdtwo{} & \(0.63\)& \(0.51\) \\
\sdxl{} & \(0.59\) & \(0.46\)\\
\sdxlturbo{} & \(0.52\)&  \(0.36\)\\
\sdxllcm{} & \(0.58\) & \(0.45\) \\
\sdvit{} & \(0.47\) & \(0.34\) \\
\fluxschnell{} & \(0.48\)& \(0.36\)\\
\fluxdev{} & \(0.47\)& \(0.32\)\\
\bottomrule
\end{tabular}
}

\caption{\textbf{GRADE score in multi- and single-prompt distributions.} All scores have a standard error of $\hat{\sigma} < 0.02$ and $\hat{\sigma} < 0.001$ respectively. We do not report standard deviation as the entropy distributions are bimodal (see \cref{fig:mean_entropy_histograms}). Values close to 1 indicate highly diverse behavior (uniform) while values close to 0 indicate highly repetitive generations. The \emph{most} diverse models are in bold.}
\label{tab:entropy_of_models}
\end{table}

\begin{figure*}[h]
    \centering
    % First Subfigure
    \begin{subfigure}[t]{0.48\textwidth} % Changed: width to 0.48\columnwidth
        \centering
        \adjustbox{max width=\linewidth, max height=5cm}{ 

            \includegraphics{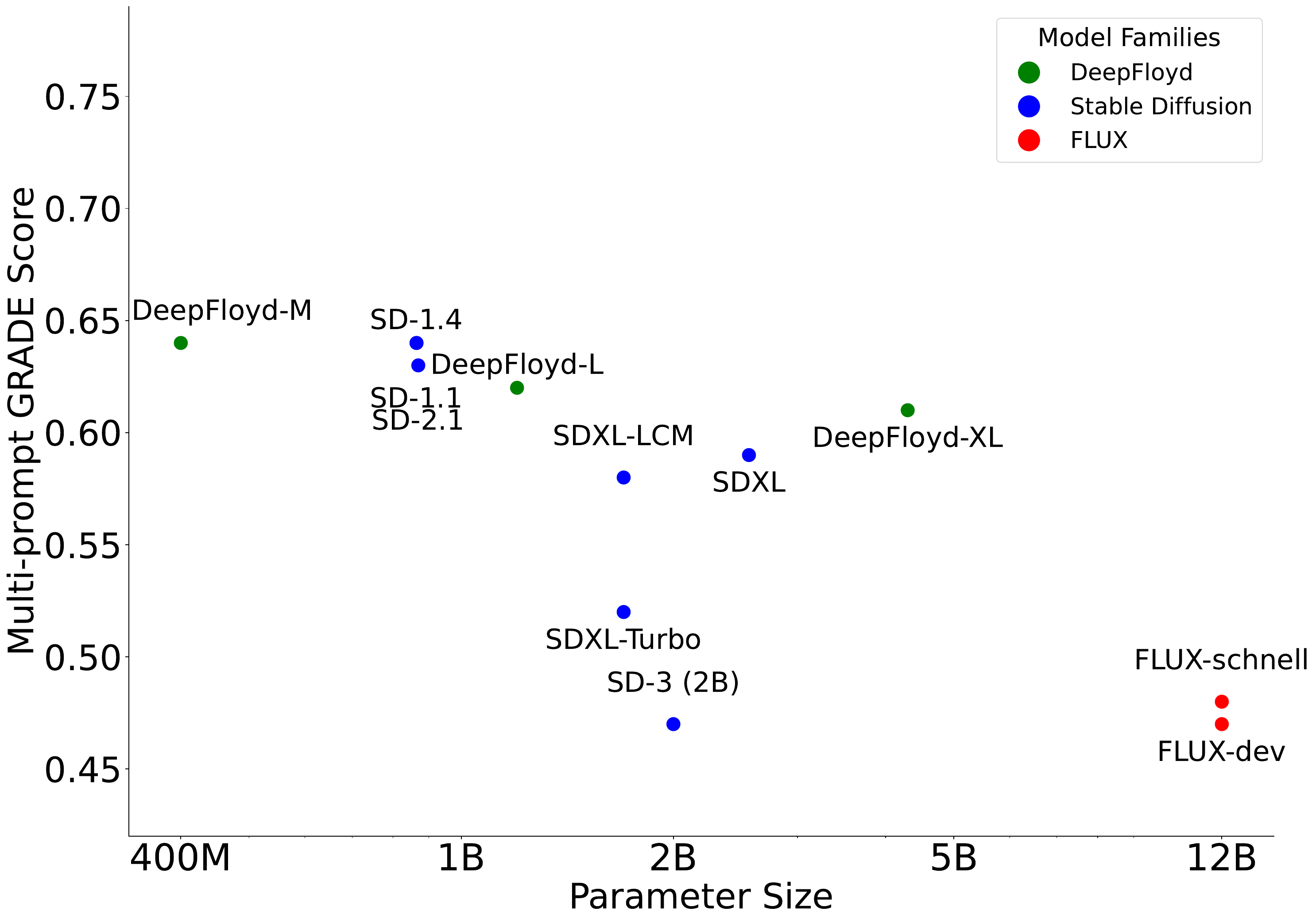}
        }
        
        % \caption{The mean concept entropy of the models plotted against the denoiser's parameter size. To a degree, diversity deteriorates in tandem with parameter size. This effect is most apparent within every model family. Models marked with $U$ denote U-Net-based models, $T$ denote transformer-based models. $U_D$ and $T_D$ denote distilled models.}
        \label{fig:inverse_scale_law}
    \end{subfigure}
    \hfill
    % Second Subfigure
    \begin{subfigure}[t]{0.48\textwidth} % Changed: width to 0.48\columnwidth
        \centering
        \adjustbox{max width=\linewidth, max height=5cm}{ 
        %     \includegraphics{Figures/concept_level_diversity_vs_pa.pdf}
        % }

        \includegraphics{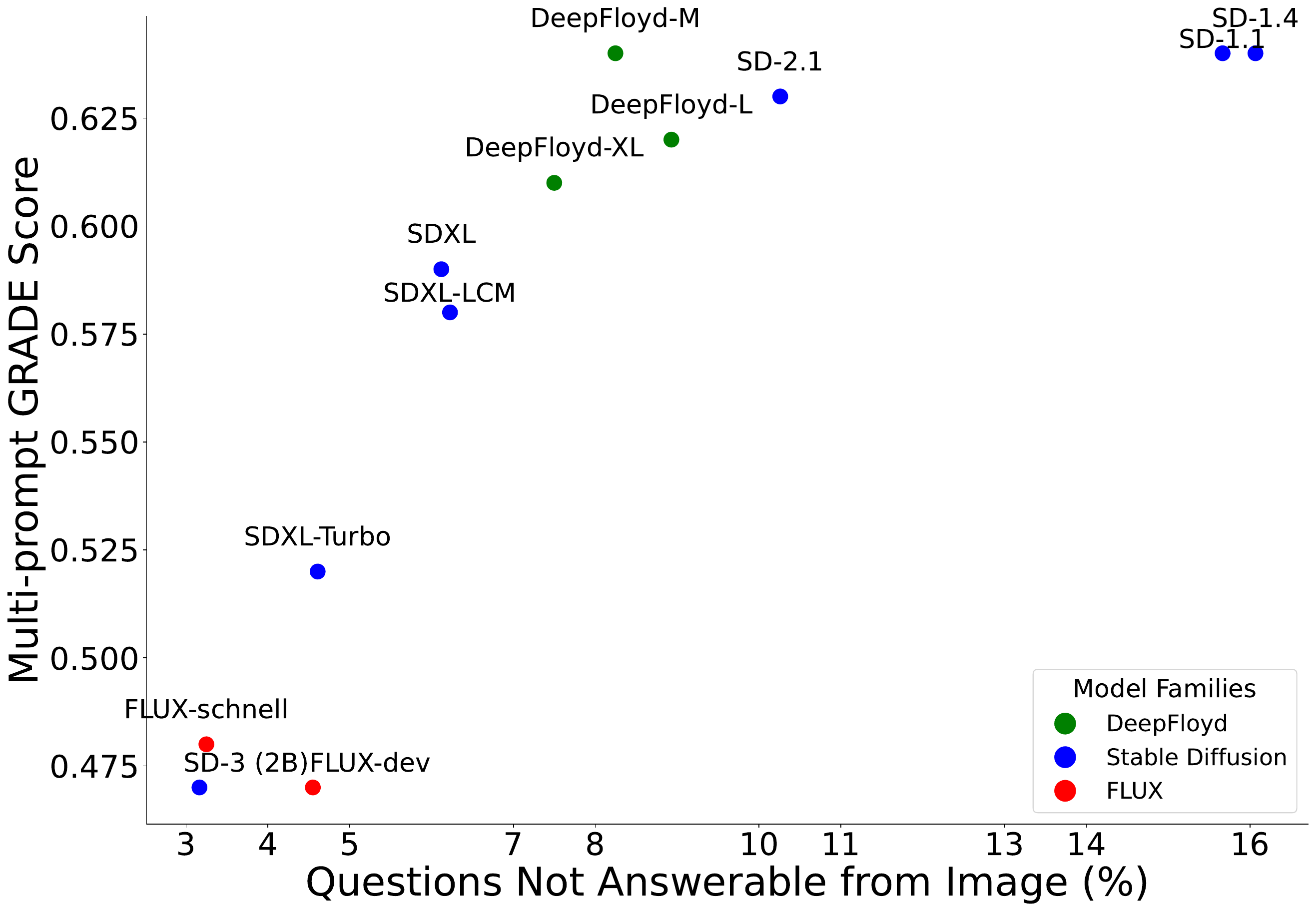}
        }

        \label{fig:diversity_vs_pa}
    \end{subfigure}
    % \caption{Comparison of mean concept entropy across different models: (a) relative to the denoiser's parameter size and (b) in relation to prompt adherence.}
    \caption{\textbf{(a) GRADE score in multi-prompt setting plotted against the denoiser's parameter size.} To a degree, diversity deteriorates in tandem with parameter size. This effect is most apparent within every model family. \textbf{(b) GRADE score in multi-prompt setting plotted against percentage of answers mapped to ``none of the above''.} In \cref{subsec:validating_grade} we show 80\% of which account for missing concepts in the image. Low ``none of the above'' values correspond to \emph{high} prompt adherence. The plot suggests a tradeoff between adherence to diversity.}

    \label{fig:combined_figures}
\end{figure*}

\textbf{All models have low diversity scores. } \cref{tab:entropy_of_models} presents the mean entropy of models across both multi- and single-prompt distributions. In \cref{app:subsec_statistical_sig_exp} we include permutation tests showing that the results are statistically significant. The average diversity across all models over multi-prompt distributions is 0.57 and 0.44 over single-prompt distributions, indicating low diversity in both categories. \cref{fig:princess_diversity_fig} illustrates the differences in diversity between models, with additional examples in \cref{app:qualitative_examples}.

% On average, multi-prompt distributions across all models have a mean entropy of 0.57, while single prompt distributions exhibit lower diversity with a mean entropy of approximately 0.44. 

\textbf{Relation of diversity to model size. } The relationship between model size and diversity suggests that diversity decreases as model size increases, as illustrated in \cref{fig:inverse_scale_law}. This trend indicates an \emph{inverse-scaling law} \citep{inverse_scale_law}, supported by Pearson $r = -0.7$ ($p = 0.011$) and Spearman $\rho = -0.84$ ($p = 0.001$) correlations between diversity and model size. However, given the small sample size of 12 models, and potential confounding factors, such as different data and architectures, we do not make any causal claims and these findings should be interpreted with caution. 
% Additionally, we do not claim that more parameters result in lower diversity due to potential confounding factors, such as different data and architectures. 
Furthermore, in addition to our claims in \cref{sec:training_data_exp} (that underspecified captions cause low diversity), \cref{fig:diversity_vs_pa} shows that the more a model generates images that are mapped to ``none of the above''\footnote{In \cref{subsec:validating_grade} we show that 80\% of unanswerable images do not depict the concept mentioned in the prompt.} (i.e., prompt adherence \emph{decreases}), the more diverse it is. Pearson $r = 0.8$ ($p = 0.02$) and Spearman $\rho = 0.94$ ($p < 0.001$) correlations reinforce this, suggesting the possibility that improving the ability of models to generate images that match the prompt is at the cost of sample diversity, similar to fidelity-diversity tradeoffs shown before \citep{classifier_guidance, improved_precision_and_recall}.

\textbf{Default behaviors. } We define \emph{default behavior} as a phenomenon where a model has a heavily skewed distribution toward a specific attribute \( \tau \geq \) \defbehthreshold{} of the time. We observe that default behaviors are highly frequent and maintain the trends in \cref{fig:combined_figures}, indicating strong correlation to entropy. All models exhibit at least one default behavior from \textbf{76\% to 90\%} of the multi-level distributions and from \textbf{87\% to 97\%} of the single prompt distributions. Similarly, the range of total default behaviors exhibited in multi-prompt distributions is between \textbf{39\% to 56\%} and between \textbf{49\% to 70\%} for single prompt distributions. Complete results with further analyses are provided in \cref{app:extended_results}.

\section{Low Diversity Originates in Training Data} 
\label{sec:training_data_exp} 

In \cref{sec:comparing_t2i_models}, we showed that T2I models often exhibit limited diversity when faced with underspecified prompts. We posit that this phenomenon stems from the nature of the training data: whenever a concept is mentioned without an explicit attribute value (e.g., ``banana'' rather than ``yellow banana''), the accompanying images in the dataset tend to be dominated by a small set of attribute values. We observe anecdotal evidence for this in LAION: sampling 100 image-caption pairs that mention a concept without specifying its attribute typically yields images that share an implicit, most common attribute value (e.g., bananas tend to be yellow). This is closely related to the linguistic phenomenon of \emph{reporting bias} \citep{reporting_bias}, where attributes deemed ``obvious'' or ``typical'' are not explicitly mentioned in captions.

Formally, each training example in a T2I dataset is a caption-image pair. When the caption includes a concept but omits an attribute (e.g., “banana”), we hypothesize that the distribution of actual images is heavily skewed toward a small subset of attribute values (most bananas in LAION are indeed yellow). As a result, the model learns to replicate this limited distribution whenever it encounters an underspecified prompt. In what follows, we verify this by comparing the distributions of (i) real images from LAION where captions omit an attribute, and (ii) generated images produced by the same underspecified captions or by similar prompts.

\subsection{Experimental Setup and Metrics}
To examine this empirically, we use \grade{} to measure diversity across multiple prompts (i.e., the \emph{multi-prompt distribution}). In particular, we measure:
\begin{itemize}
    \item \textbf{Training data distribution:} We select underspecified captions from LAION (e.g., ``cookie'' but not ``cookie cutter,'' and with no mention or implication of a specific attribute such as shape). We refer to these as \emph{filtered captions}. 
    \item \textbf{Model-generated distribution:} We use the same underspecified captions (and also additional unseen prompts) as inputs to a T2I model and generate multiple images per prompt. We then measure the distribution of attribute values across these generated images.
\end{itemize}
We compare these distributions using three statistics: (1) \emph{entropy}, which captures overall diversity; (2) Pearson correlation coefficient (PCC), which captures the extent to which the attribute-value frequencies align between the training data and the generated images; and (3) TVD, which measures the dissimilarity between the two distributions.

\textbf{Replication of training data diversity.}
First, we test whether T2I models reproduce the diversity observed in underspecified caption-image pairs from their own training data. For each model, we select 50 triplets of concepts, attributes, and attribute values. We filter LAION captions that mention the concept as an object but do not specify or imply the attribute (e.g., ``a cookie on a table'' as opposed to ``a classic chocolate chip cookie,'' which implies it is round). We then:
\begin{enumerate}
    \item Collect up to 150 such \emph{filtered captions} per concept from LAION (technical details are presented in \cref{app:subsec:laion_experiment}).
    \item Compute \grade{} on the actual images associated with these captions.
    \item Generate 20 images per filtered caption using a T2I model, thus obtaining 3,000 generated images per concept.
    \item Compute \grade{} on these generated images to obtain their distribution of attribute values.
    \item Compare the real (LAION) and generated distributions via entropy, PCC, and TVD.
\end{enumerate}

% \textbf{Generalizing to new underspecified prompts.}
% We next explore whether the model’s tendency to mirror training data extends to prompts that are not sampled from LAION. Specifically, we use the multi-prompt distributions obtained in \cref{sec:comparing_t2i_models}. We then compare them with the corresponding distributions from LAION for the same concept-attribute pairs. This comparison reveals whether the model continues to replicate the underspecified distributions it observed in the training set.

\textbf{Generalizing to new underspecified prompts.}
We next explore whether the model’s tendency to mirror training data extends to prompts that are not sampled from LAION. Specifically, we compare the multi-prompt distributions obtained in \cref{sec:comparing_t2i_models} with the corresponding distributions from LAION for the same concept-attribute pairs. This comparison reveals whether the model continues to replicate the underspecified distributions it observed in the training set.

\subsection{Results}
\label{subsec:training_data_results}
\Cref{tab:comparison_to_laion_exp} summarizes the outcomes. LAION itself exhibits moderate diversity for our selected concepts, reflected by dataset entropy values of 0.64 in \laiontwob{} and 0.65 in \laionfiveb{}. When prompted with the \emph{exact filtered captions} from LAION, models achieve a similar range of entropy (0.62--0.68). The correlation between model outputs and LAION images is high (PCC of 0.73--0.88), and the TVD remains low (0.10--0.13). These observations imply that T2I models replicate the underspecified distributions seen in their own training data.

When the same models are provided with \emph{new}, underspecified prompts (``Generated'' in \cref{tab:comparison_to_laion_exp}), the alignment with LAION images diminishes slightly. The PCC drops (0.61--0.72 vs.\ 0.73--0.88), and the TVD increases marginally (0.17--0.18 vs.\ 0.10--0.13). Yet, the overall trend remains the same: the generated multi-prompt distributions still resemble those in LAION for the given concept-attribute pairs.

These results strongly support our core hypothesis: when concept-attribute pairs are left unspecified in captions, most images in the training data depict a single implicit, most common attribute value. Consequently, T2I models learn to replicate this bias. Unless the user explicitly overrides it with a specific attribute, the model reproduces the distribution it has observed most frequently in training, resulting in a systematic lack of diversity.

\begin{table}[t]
\centering
\resizebox{\columnwidth}{!}{%
\begin{tabular}{lllcccc}
\toprule
\textbf{Model} & \textbf{Dataset} & \textbf{Source of Prompts} & \multicolumn{2}{c}{\textbf{Entropy}} & \multicolumn{2}{c}{\textbf{Similarity}} \\
\cmidrule(lr){4-5} \cmidrule(lr){6-7}
               &                  &                          & \textbf{Model} & \textbf{Dataset} & \textbf{PCC} & \textbf{TVD} \\
\midrule
\multirow{2}{*}{\sdclassicfirstckpt{}} 
    & \multirow{2}{*}{\laiontwob{}} 
    & \laiontwob{} & $0.62$ & \multirow{2}{*}{$0.64$} & $0.86$ & $0.11$ \\
    &                             
    & Generated     & $0.58$ &                    & $0.71$ & $0.18$ \\
\midrule
\multirow{2}{*}{\sdclassic{}}        
    & \multirow{2}{*}{\laiontwob{}} 
    & \laiontwob{} & $0.62$ & \multirow{2}{*}{$0.64$} & $0.88$ & $0.10$ \\
    &                             
    & Generated     & $0.60$ &                    & $0.72$ & $0.17$ \\
\midrule
\multirow{2}{*}{\sdtwo{}}            
    & \multirow{2}{*}{\laionfiveb{}} 
    & \laionfiveb{} & $0.68$ & \multirow{2}{*}{$0.65$} & $0.73$ & $0.13$ \\
    &                             
    & Generated     & $0.68$ &                    & $0.61$ & $0.18$ \\
\bottomrule
\end{tabular}
}
\caption{\textbf{Similarities between model outputs and its training set.} The entropy values, PCC, and TVD all indicate models have comparable diversity to the training set.}
\label{tab:comparison_to_laion_exp}
\end{table}

% \section{Limitations} \grade{} has two main limitations. First, as any diversity measurement, its scores only reflect the diversity of concepts that were measured. Moreover, since it assesses diversity across specific attributes, it does not  capture unmeasured ones. Second, \grade{} depends on its underlying LLM and VQA, whose unknown biases may affect both the LLM's suggestions, the quality of the VQA-extracted information, and the final diversity score. 

\section{Limitations}
While \grade{} provides a fine-grained view of sample diversity, it has two limitations. First, as with any metric focused on a specific set of concepts and attributes, its scores depend heavily on which attributes are measured; attributes not included in the evaluation remain unassessed. Second, it relies on external LLM and VQA components, introducing potential biases and inaccuracies from these models into both the attribute-suggestion process and the final diversity score. Despite these limitations, we believe that \grade{} represents a step toward more interpretable, fine-grained diversity assessments in T2I models.
\section{Conclusion}\label{sec:conclusion}
% We introduce \grade{}, an automatic, fine-grained method for measuring sample diversity in T2I models based on concepts and their attributes. By estimating the distribution of attributes of a concept in generated images, \grade{} provides a diversity score that can be used to interpret model behavior. Unlike traditional diversity metrics, \grade{} does not rely on reference images. 

% Our experiments demonstrate that humans find \grade{} accurate, while at the same time showing weak correlation with traditional metrics like \fid{} and \recall{}. We compare the diversity of \numModelsTested{} models and find that current models generate semantically repetitive images across different prompts on anywhere from 78\% to 90\% of the concepts tested, with an increasing trend as models scale in size and improve in prompt adherence, highlighting a limited ability to capture the rich diversity inherent in visual concepts. We further hypothesize that diversity in generation is linked to diversity in the training data and that underspecification encourages default behavior.

% Future work could explore methods to enrich training data, incorporate diversity-promoting mechanisms during model training, and extend \grade{} to evaluate relationships between different attributes of a concept or the relationship between multiple concepts in a scene. 

% Below is a more concise conclusion that highlights GRADE’s advantages over existing metrics and encourages its use:

We presented GRADE, a reference-free and fine-grained approach for measuring semantic diversity in T2I models. Unlike traditional metrics that rely on global distribution comparisons (e.g., FID or Precision-Recall), GRADE focuses on concept-specific attributes, providing an interpretable view into how consistently models capture the variety of real-world concepts. Our experiments show that current T2I models—regardless of parameter size—often converge on default attributes and produce semantically repetitive images, revealing a concerning lack of diversity. Notably, larger models yield less varied outputs, hinting at an inverse-scaling trend that underscores the need to address underspecified training data and design objectives that explicitly foster diversity.

By leveraging an LLM and a VQA system, GRADE automates diversity analysis with minimal overhead and no reliance on curated reference datasets. We encourage researchers to adopt GRADE not only for diagnosing a model’s limitations but also for guiding future refinements—such as improving training data quality or designing diversity-driven model objectives. Exploring multi-attribute relationships, combining GRADE with other evaluation measures, and investigating training interventions are promising directions for further work. Ultimately, we believe that GRADE can push the field toward developing T2I systems that capture the true breadth and richness of visual concepts.

%We hope our work will inspire more nuanced evaluations and drive advancements in generating diverse visual content from textual descriptions.

% conduct a comprehensive analysis of \numModelsTested{} state-of-the-art T2I models and uncovered a prevalent limitation: these models default to generating images with the same attributes for a concept on anywhere from 78\% to 90\% of the concepts we tested, with an increasing trend as models scale in size and improve in prompt adherence, highlighting a limited ability to capture the rich diversity inherent in visual concepts. We further hypothesize that diversity in generation is linked to diversity in the training data, and that under-specification encourages default behavior.
% Future work could explore methods to enrich training data, incorporate diversity-promoting mechanisms during model training, and extend \grade{} to evaluate relationships between different attributes of a concept or the relationship between multiple concepts in a scene.  We hope our work will inspire more nuanced evaluations and drive advancements in generating diverse visual content from textual descriptions.

% \section{Final copy}
% \royi{do not forget to include}
% You must include your signed IEEE copyright release form when you submit your finished paper.
% \textbf{We MUST have this form before your paper can be published in the proceedings.
% }
% Please direct any questions to the production editor in charge of these proceedings at the IEEE Computer Society Press:
% \url{https://www.computer.org/about/contact}.
\newpage
\clearpage
{
    \small
    \bibliographystyle{ieeenat_fullname}
    \bibliography{main}

\begin{thebibliography}{40}
\providecommand{\natexlab}[1]{#1}
\providecommand{\url}[1]{\texttt{#1}}
\expandafter\ifx\csname urlstyle\endcsname\relax
  \providecommand{\doi}[1]{doi: #1}\else
  \providecommand{\doi}{doi: \begingroup \urlstyle{rm}\Url}\fi

\bibitem[Alaa et~al.(2022)Alaa, Van~Breugel, Saveliev, and van~der Schaar]{faithful_synthetic_data}
Ahmed Alaa, Boris Van~Breugel, Evgeny~S Saveliev, and Mihaela van~der Schaar.
\newblock How faithful is your synthetic data? sample-level metrics for evaluating and auditing generative models.
\newblock In \emph{International Conference on Machine Learning}, pages 290--306. PMLR, 2022.

\bibitem[at~StabilityAI(2023)]{DeepFloydIF}
DeepFloyd~Lab at StabilityAI.
\newblock {DeepFloyd IF}: a novel state-of-the-art open-source text-to-image model with a high degree of photorealism and language understanding.
\newblock \url{https://www.deepfloyd.ai/deepfloyd-if}, 2023.
\newblock Retrieved on 2023-11-08.

\bibitem[Bi{\'n}kowski et~al.(2018)Bi{\'n}kowski, Sutherland, Arbel, and Gretton]{fid_sample_size_bias}
Miko{\l}aj Bi{\'n}kowski, Danica~J Sutherland, Michael Arbel, and Arthur Gretton.
\newblock Demystifying mmd gans.
\newblock \emph{arXiv preprint arXiv:1801.01401}, 2018.

\bibitem[{Black Forest Labs}(2024{\natexlab{a}})]{blackforestlabs2024dev}
{Black Forest Labs}.
\newblock {FLUX.1-dev Model Documentation}.
\newblock \url{https://huggingface.co/black-forest-labs/FLUX.1-dev}, 2024{\natexlab{a}}.
\newblock Accessed: Aug 24 2024.

\bibitem[{Black Forest Labs}(2024{\natexlab{b}})]{blackforestlabs2024schnell}
{Black Forest Labs}.
\newblock {FLUX.1-dev Model Documentation}.
\newblock \url{https://huggingface.co/black-forest-labs/FLUX.1-schnell}, 2024{\natexlab{b}}.
\newblock Accessed: Aug 24 2024.

\bibitem[Bonnini et~al.(2024)Bonnini, Assegie, Kamila, et~al.]{bonnini2024review}
Stefano Bonnini, Getnet~Melak Assegie, Trzcinska Kamila, et~al.
\newblock Review about the permutation approach in hypothesis testing.
\newblock \emph{Mathematics}, 12:\penalty0 2617--1, 2024.

\bibitem[Chong and Forsyth(2020)]{fid_biased_to_bias}
Min~Jin Chong and David Forsyth.
\newblock Effectively unbiased fid and inception score and where to find them.
\newblock In \emph{Proceedings of the IEEE/CVF conference on computer vision and pattern recognition}, pages 6070--6079, 2020.

\bibitem[Deng et~al.(2009)Deng, Dong, Socher, Li, Li, and Fei-Fei]{imagenet}
Jia Deng, Wei Dong, Richard Socher, Li-Jia Li, Kai Li, and Li Fei-Fei.
\newblock Imagenet: A large-scale hierarchical image database.
\newblock In \emph{2009 IEEE conference on computer vision and pattern recognition}, pages 248--255. Ieee, 2009.

\bibitem[Dhariwal and Nichol(2021)]{classifier_guidance}
Prafulla Dhariwal and Alexander Nichol.
\newblock Diffusion models beat gans on image synthesis.
\newblock \emph{Advances in neural information processing systems}, 34:\penalty0 8780--8794, 2021.

\bibitem[D'Inc{\`a} et~al.(2024)D'Inc{\`a}, Peruzzo, Mancini, Xu, Goel, Xu, Wang, Shi, and Sebe]{openbias}
Moreno D'Inc{\`a}, Elia Peruzzo, Massimiliano Mancini, Dejia Xu, Vidit Goel, Xingqian Xu, Zhangyang Wang, Humphrey Shi, and Nicu Sebe.
\newblock Openbias: Open-set bias detection in text-to-image generative models.
\newblock In \emph{Proceedings of the IEEE/CVF Conference on Computer Vision and Pattern Recognition}, pages 12225--12235, 2024.

\bibitem[Elazar et~al.(2024)Elazar, Bhagia, Magnusson, Ravichander, Schwenk, Suhr, Walsh, Groeneveld, Soldaini, Singh, Hajishirzi, Smith, and Dodge]{wimbd}
Yanai Elazar, Akshita Bhagia, Ian Magnusson, Abhilasha Ravichander, Dustin Schwenk, Alane Suhr, Pete Walsh, Dirk Groeneveld, Luca Soldaini, Sameer Singh, Hanna Hajishirzi, Noah~A. Smith, and Jesse Dodge.
\newblock What's in my big data?, 2024.

\bibitem[Esser et~al.(2024)Esser, Kulal, Blattmann, Entezari, M{\"u}ller, Saini, Levi, Lorenz, Sauer, Boesel, et~al.]{stable_diffusion_vit}
Patrick Esser, Sumith Kulal, Andreas Blattmann, Rahim Entezari, Jonas M{\"u}ller, Harry Saini, Yam Levi, Dominik Lorenz, Axel Sauer, Frederic Boesel, et~al.
\newblock Scaling rectified flow transformers for high-resolution image synthesis.
\newblock \emph{arXiv preprint arXiv:2403.03206}, 2024.

\bibitem[Friedman and Dieng(2022)]{vendiscore}
Dan Friedman and Adji~Bousso Dieng.
\newblock The vendi score: A diversity evaluation metric for machine learning.
\newblock \emph{arXiv preprint arXiv:2210.02410}, 2022.

\bibitem[Gordon and Van~Durme(2013)]{reporting_bias}
Jonathan Gordon and Benjamin Van~Durme.
\newblock Reporting bias and knowledge acquisition.
\newblock In \emph{Proceedings of the 2013 Workshop on Automated Knowledge Base Construction}, page 25–30, New York, NY, USA, 2013. Association for Computing Machinery.

\bibitem[Heusel et~al.(2017)Heusel, Ramsauer, Unterthiner, Nessler, and Hochreiter]{fid_paper}
Martin Heusel, Hubert Ramsauer, Thomas Unterthiner, Bernhard Nessler, and Sepp Hochreiter.
\newblock Gans trained by a two time-scale update rule converge to a local nash equilibrium.
\newblock In \emph{Advances in Neural Information Processing Systems}. Curran Associates, Inc., 2017.

\bibitem[Hutchinson et~al.(2022)Hutchinson, Baldridge, and Prabhakaran]{underspecified_t2i}
Ben Hutchinson, Jason Baldridge, and Vinodkumar Prabhakaran.
\newblock Underspecification in scene description-to-depiction tasks.
\newblock \emph{arXiv preprint arXiv:2210.05815}, 2022.

\bibitem[Ilharco et~al.(2021)Ilharco, Wortsman, Wightman, Gordon, Carlini, Taori, Dave, Shankar, Namkoong, Miller, Hajishirzi, Farhadi, and Schmidt]{openai_clip_software}
Gabriel Ilharco, Mitchell Wortsman, Ross Wightman, Cade Gordon, Nicholas Carlini, Rohan Taori, Achal Dave, Vaishaal Shankar, Hongseok Namkoong, John Miller, Hannaneh Hajishirzi, Ali Farhadi, and Ludwig Schmidt.
\newblock Openclip, 2021.
\newblock If you use this software, please cite it as below.

\bibitem[Jayasumana et~al.(2024)Jayasumana, Ramalingam, Veit, Glasner, Chakrabarti, and Kumar]{rethinking_fid}
Sadeep Jayasumana, Srikumar Ramalingam, Andreas Veit, Daniel Glasner, Ayan Chakrabarti, and Sanjiv Kumar.
\newblock Rethinking fid: Towards a better evaluation metric for image generation.
\newblock In \emph{Proceedings of the IEEE/CVF Conference on Computer Vision and Pattern Recognition}, pages 9307--9315, 2024.

\bibitem[Kim et~al.(2023)Kim, Kwon, and Uh]{attribute_based_diversity}
Dongkyun Kim, Mingi Kwon, and Youngjung Uh.
\newblock Attribute based interpretable evaluation metrics for generative models.
\newblock \emph{arXiv preprint arXiv:2310.17261}, 2023.

\bibitem[Kynk{\"a}{\"a}nniemi et~al.(2019)Kynk{\"a}{\"a}nniemi, Karras, Laine, Lehtinen, and Aila]{improved_precision_and_recall}
Tuomas Kynk{\"a}{\"a}nniemi, Tero Karras, Samuli Laine, Jaakko Lehtinen, and Timo Aila.
\newblock Improved precision and recall metric for assessing generative models.
\newblock \emph{Advances in neural information processing systems}, 32, 2019.

\bibitem[Kynk{\"a}{\"a}nniemi et~al.(2022)Kynk{\"a}{\"a}nniemi, Karras, Aittala, Aila, and Lehtinen]{fid_imagenet_bias}
Tuomas Kynk{\"a}{\"a}nniemi, Tero Karras, Miika Aittala, Timo Aila, and Jaakko Lehtinen.
\newblock The role of imagenet classes in fr$\backslash$'echet inception distance.
\newblock \emph{arXiv preprint arXiv:2203.06026}, 2022.

\bibitem[Labs(2024)]{flux_announcement}
Black~Forest Labs.
\newblock Announcing black forest labs.
\newblock \url{https://blackforestlabs.ai/announcing-black-forest-labs/}, 2024.
\newblock Accessed: 2024-08-29.

\bibitem[Luo et~al.(2023)Luo, Tan, Patil, Gu, von Platen, Passos, Huang, Li, and Zhao]{sdxl_lcm}
Simian Luo, Yiqin Tan, Suraj Patil, Daniel Gu, Patrick von Platen, Apolin{\'a}rio Passos, Longbo Huang, Jian Li, and Hang Zhao.
\newblock Lcm-lora: A universal stable-diffusion acceleration module.
\newblock \emph{arXiv preprint arXiv:2311.05556}, 2023.

\bibitem[McKenzie et~al.(2023)McKenzie, Lyzhov, Pieler, Parrish, Mueller, Prabhu, McLean, Kirtland, Ross, Liu, et~al.]{inverse_scale_law}
Ian~R McKenzie, Alexander Lyzhov, Michael Pieler, Alicia Parrish, Aaron Mueller, Ameya Prabhu, Euan McLean, Aaron Kirtland, Alexis Ross, Alisa Liu, et~al.
\newblock Inverse scaling: When bigger isn't better.
\newblock \emph{arXiv preprint arXiv:2306.09479}, 2023.

\bibitem[Naeem et~al.(2020)Naeem, Oh, Uh, Choi, and Yoo]{density_and_coverage}
Muhammad~Ferjad Naeem, Seong~Joon Oh, Youngjung Uh, Yunjey Choi, and Jaejun Yoo.
\newblock Reliable fidelity and diversity metrics for generative models.
\newblock In \emph{International Conference on Machine Learning}, pages 7176--7185. PMLR, 2020.

\bibitem[OpenAI(2024)]{openai_structured_outputs_api}
OpenAI.
\newblock Introducing structured outputs in the api, 2024.
\newblock Accessed: 2024-09-17.

\bibitem[OpenAI et~al.(2024)OpenAI, Achiam, Adler, Agarwal, Ahmad, Akkaya, Aleman, Almeida, Altenschmidt, Altman, Anadkat, Avila, Babuschkin, Balaji, Balcom, Baltescu, Bao, Bavarian, Belgum, Bello, Berdine, Bernadett-Shapiro, Berner, Bogdonoff, Boiko, Boyd, Brakman, Brockman, Brooks, Brundage, Button, Cai, Campbell, Cann, Carey, Carlson, Carmichael, Chan, Chang, Chantzis, Chen, Chen, Chen, Chen, Chen, Chess, Cho, Chu, Chung, Cummings, Currier, Dai, Decareaux, Degry, Deutsch, Deville, Dhar, Dohan, Dowling, Dunning, Ecoffet, Eleti, Eloundou, Farhi, Fedus, Felix, Fishman, Forte, Fulford, Gao, Georges, Gibson, Goel, Gogineni, Goh, Gontijo-Lopes, Gordon, Grafstein, Gray, Greene, Gross, Gu, Guo, Hallacy, Han, Harris, He, Heaton, Heidecke, Hesse, Hickey, Hickey, Hoeschele, Houghton, Hsu, Hu, Hu, Huizinga, Jain, Jain, Jang, Jiang, Jiang, Jin, Jin, Jomoto, Jonn, Jun, Kaftan, Łukasz Kaiser, Kamali, Kanitscheider, Keskar, Khan, Kilpatrick, Kim, Kim, Kim, Kirchner, Kiros, Knight, Kokotajlo, Łukasz Kondraciuk, Kondrich,
  Konstantinidis, Kosic, Krueger, Kuo, Lampe, Lan, Lee, Leike, Leung, Levy, Li, Lim, Lin, Lin, Litwin, Lopez, Lowe, Lue, Makanju, Malfacini, Manning, Markov, Markovski, Martin, Mayer, Mayne, McGrew, McKinney, McLeavey, McMillan, McNeil, Medina, Mehta, Menick, Metz, Mishchenko, Mishkin, Monaco, Morikawa, Mossing, Mu, Murati, Murk, Mély, Nair, Nakano, Nayak, Neelakantan, Ngo, Noh, Ouyang, O'Keefe, Pachocki, Paino, Palermo, Pantuliano, Parascandolo, Parish, Parparita, Passos, Pavlov, Peng, Perelman, de~Avila Belbute~Peres, Petrov, de~Oliveira~Pinto, Michael, Pokorny, Pokrass, Pong, Powell, Power, Power, Proehl, Puri, Radford, Rae, Ramesh, Raymond, Real, Rimbach, Ross, Rotsted, Roussez, Ryder, Saltarelli, Sanders, Santurkar, Sastry, Schmidt, Schnurr, Schulman, Selsam, Sheppard, Sherbakov, Shieh, Shoker, Shyam, Sidor, Sigler, Simens, Sitkin, Slama, Sohl, Sokolowsky, Song, Staudacher, Such, Summers, Sutskever, Tang, Tezak, Thompson, Tillet, Tootoonchian, Tseng, Tuggle, Turley, Tworek, Uribe, Vallone, Vijayvergiya,
  Voss, Wainwright, Wang, Wang, Wang, Ward, Wei, Weinmann, Welihinda, Welinder, Weng, Weng, Wiethoff, Willner, Winter, Wolrich, Wong, Workman, Wu, Wu, Wu, Xiao, Xu, Yoo, Yu, Yuan, Zaremba, Zellers, Zhang, Zhang, Zhao, Zheng, Zhuang, Zhuk, and Zoph]{gpt4}
OpenAI, Josh Achiam, Steven Adler, Sandhini Agarwal, Lama Ahmad, Ilge Akkaya, Florencia~Leoni Aleman, Diogo Almeida, Janko Altenschmidt, Sam Altman, Shyamal Anadkat, Red Avila, Igor Babuschkin, Suchir Balaji, Valerie Balcom, Paul Baltescu, Haiming Bao, Mohammad Bavarian, Jeff Belgum, Irwan Bello, Jake Berdine, Gabriel Bernadett-Shapiro, Christopher Berner, Lenny Bogdonoff, Oleg Boiko, Madelaine Boyd, Anna-Luisa Brakman, Greg Brockman, Tim Brooks, Miles Brundage, Kevin Button, Trevor Cai, Rosie Campbell, Andrew Cann, Brittany Carey, Chelsea Carlson, Rory Carmichael, Brooke Chan, Che Chang, Fotis Chantzis, Derek Chen, Sully Chen, Ruby Chen, Jason Chen, Mark Chen, Ben Chess, Chester Cho, Casey Chu, Hyung~Won Chung, Dave Cummings, Jeremiah Currier, Yunxing Dai, Cory Decareaux, Thomas Degry, Noah Deutsch, Damien Deville, Arka Dhar, David Dohan, Steve Dowling, Sheila Dunning, Adrien Ecoffet, Atty Eleti, Tyna Eloundou, David Farhi, Liam Fedus, Niko Felix, Simón~Posada Fishman, Juston Forte, Isabella Fulford, Leo
  Gao, Elie Georges, Christian Gibson, Vik Goel, Tarun Gogineni, Gabriel Goh, Rapha Gontijo-Lopes, Jonathan Gordon, Morgan Grafstein, Scott Gray, Ryan Greene, Joshua Gross, Shixiang~Shane Gu, Yufei Guo, Chris Hallacy, Jesse Han, Jeff Harris, Yuchen He, Mike Heaton, Johannes Heidecke, Chris Hesse, Alan Hickey, Wade Hickey, Peter Hoeschele, Brandon Houghton, Kenny Hsu, Shengli Hu, Xin Hu, Joost Huizinga, Shantanu Jain, Shawn Jain, Joanne Jang, Angela Jiang, Roger Jiang, Haozhun Jin, Denny Jin, Shino Jomoto, Billie Jonn, Heewoo Jun, Tomer Kaftan, Łukasz Kaiser, Ali Kamali, Ingmar Kanitscheider, Nitish~Shirish Keskar, Tabarak Khan, Logan Kilpatrick, Jong~Wook Kim, Christina Kim, Yongjik Kim, Jan~Hendrik Kirchner, Jamie Kiros, Matt Knight, Daniel Kokotajlo, Łukasz Kondraciuk, Andrew Kondrich, Aris Konstantinidis, Kyle Kosic, Gretchen Krueger, Vishal Kuo, Michael Lampe, Ikai Lan, Teddy Lee, Jan Leike, Jade Leung, Daniel Levy, Chak~Ming Li, Rachel Lim, Molly Lin, Stephanie Lin, Mateusz Litwin, Theresa Lopez, Ryan
  Lowe, Patricia Lue, Anna Makanju, Kim Malfacini, Sam Manning, Todor Markov, Yaniv Markovski, Bianca Martin, Katie Mayer, Andrew Mayne, Bob McGrew, Scott~Mayer McKinney, Christine McLeavey, Paul McMillan, Jake McNeil, David Medina, Aalok Mehta, Jacob Menick, Luke Metz, Andrey Mishchenko, Pamela Mishkin, Vinnie Monaco, Evan Morikawa, Daniel Mossing, Tong Mu, Mira Murati, Oleg Murk, David Mély, Ashvin Nair, Reiichiro Nakano, Rajeev Nayak, Arvind Neelakantan, Richard Ngo, Hyeonwoo Noh, Long Ouyang, Cullen O'Keefe, Jakub Pachocki, Alex Paino, Joe Palermo, Ashley Pantuliano, Giambattista Parascandolo, Joel Parish, Emy Parparita, Alex Passos, Mikhail Pavlov, Andrew Peng, Adam Perelman, Filipe de Avila Belbute~Peres, Michael Petrov, Henrique~Ponde de Oliveira~Pinto, Michael, Pokorny, Michelle Pokrass, Vitchyr~H. Pong, Tolly Powell, Alethea Power, Boris Power, Elizabeth Proehl, Raul Puri, Alec Radford, Jack Rae, Aditya Ramesh, Cameron Raymond, Francis Real, Kendra Rimbach, Carl Ross, Bob Rotsted, Henri Roussez,
  Nick Ryder, Mario Saltarelli, Ted Sanders, Shibani Santurkar, Girish Sastry, Heather Schmidt, David Schnurr, John Schulman, Daniel Selsam, Kyla Sheppard, Toki Sherbakov, Jessica Shieh, Sarah Shoker, Pranav Shyam, Szymon Sidor, Eric Sigler, Maddie Simens, Jordan Sitkin, Katarina Slama, Ian Sohl, Benjamin Sokolowsky, Yang Song, Natalie Staudacher, Felipe~Petroski Such, Natalie Summers, Ilya Sutskever, Jie Tang, Nikolas Tezak, Madeleine~B. Thompson, Phil Tillet, Amin Tootoonchian, Elizabeth Tseng, Preston Tuggle, Nick Turley, Jerry Tworek, Juan Felipe~Cerón Uribe, Andrea Vallone, Arun Vijayvergiya, Chelsea Voss, Carroll Wainwright, Justin~Jay Wang, Alvin Wang, Ben Wang, Jonathan Ward, Jason Wei, CJ Weinmann, Akila Welihinda, Peter Welinder, Jiayi Weng, Lilian Weng, Matt Wiethoff, Dave Willner, Clemens Winter, Samuel Wolrich, Hannah Wong, Lauren Workman, Sherwin Wu, Jeff Wu, Michael Wu, Kai Xiao, Tao Xu, Sarah Yoo, Kevin Yu, Qiming Yuan, Wojciech Zaremba, Rowan Zellers, Chong Zhang, Marvin Zhang, Shengjia
  Zhao, Tianhao Zheng, Juntang Zhuang, William Zhuk, and Barret Zoph.
\newblock Gpt-4 technical report, 2024.

\bibitem[Parmar et~al.(2022)Parmar, Zhang, and Zhu]{fid_image_processing}
Gaurav Parmar, Richard Zhang, and Jun-Yan Zhu.
\newblock On aliased resizing and surprising subtleties in gan evaluation.
\newblock In \emph{Proceedings of the IEEE/CVF Conference on Computer Vision and Pattern Recognition}, pages 11410--11420, 2022.

\bibitem[Pasarkar and Dieng(2023)]{cousins_of_vendiscore}
Amey~P Pasarkar and Adji~Bousso Dieng.
\newblock Cousins of the vendi score: A family of similarity-based diversity metrics for science and machine learning.
\newblock \emph{arXiv preprint arXiv:2310.12952}, 2023.

\bibitem[Podell et~al.(2023)Podell, English, Lacey, Blattmann, Dockhorn, M{\"u}ller, Penna, and Rombach]{sdxl}
Dustin Podell, Zion English, Kyle Lacey, Andreas Blattmann, Tim Dockhorn, Jonas M{\"u}ller, Joe Penna, and Robin Rombach.
\newblock Sdxl: Improving latent diffusion models for high-resolution image synthesis.
\newblock \emph{arXiv preprint arXiv:2307.01952}, 2023.

\bibitem[Radford et~al.(2021)Radford, Kim, Hallacy, Ramesh, Goh, Agarwal, Sastry, Askell, Mishkin, Clark, et~al.]{clipscore}
Alec Radford, Jong~Wook Kim, Chris Hallacy, Aditya Ramesh, Gabriel Goh, Sandhini Agarwal, Girish Sastry, Amanda Askell, Pamela Mishkin, Jack Clark, et~al.
\newblock Learning transferable visual models from natural language supervision.
\newblock In \emph{International conference on machine learning}, pages 8748--8763. PMLR, 2021.

\bibitem[Rassin et~al.(2022)Rassin, Ravfogel, and Goldberg]{dalle_seeing_double}
Royi Rassin, Shauli Ravfogel, and Yoav Goldberg.
\newblock Dalle-2 is seeing double: flaws in word-to-concept mapping in text2image models.
\newblock \emph{arXiv preprint arXiv:2210.10606}, 2022.

\bibitem[Rombach et~al.(2022)Rombach, Blattmann, Lorenz, Esser, and Ommer]{stable_diffusion}
Robin Rombach, Andreas Blattmann, Dominik Lorenz, Patrick Esser, and Bj\"orn Ommer.
\newblock High-resolution image synthesis with latent diffusion models.
\newblock In \emph{Proceedings of the IEEE/CVF Conference on Computer Vision and Pattern Recognition (CVPR)}, pages 10684--10695, 2022.

\bibitem[Sajjadi et~al.(2018)Sajjadi, Bachem, Lucic, Bousquet, and Gelly]{precision_and_recall}
Mehdi~SM Sajjadi, Olivier Bachem, Mario Lucic, Olivier Bousquet, and Sylvain Gelly.
\newblock Assessing generative models via precision and recall.
\newblock \emph{Advances in neural information processing systems}, 31, 2018.

\bibitem[Salimans et~al.(2016)Salimans, Goodfellow, Zaremba, Cheung, Radford, and Chen]{inception_score}
Tim Salimans, Ian Goodfellow, Wojciech Zaremba, Vicki Cheung, Alec Radford, and Xi Chen.
\newblock Improved techniques for training gans.
\newblock \emph{Advances in neural information processing systems}, 29, 2016.

\bibitem[Sauer et~al.(2023)Sauer, Lorenz, Blattmann, and Rombach]{sdxl_turbo}
Axel Sauer, Dominik Lorenz, Andreas Blattmann, and Robin Rombach.
\newblock Adversarial diffusion distillation.
\newblock \emph{arXiv preprint arXiv:2311.17042}, 2023.

\bibitem[Schuhmann et~al.(2022)Schuhmann, Beaumont, Vencu, Gordon, Wightman, Cherti, Coombes, Katta, Mullis, Wortsman, Schramowski, Kundurthy, Crowson, Schmidt, Kaczmarczyk, and Jitsev]{laion-5b-paper}
Christoph Schuhmann, Romain Beaumont, Richard Vencu, Cade~W Gordon, Ross Wightman, Mehdi Cherti, Theo Coombes, Aarush Katta, Clayton Mullis, Mitchell Wortsman, Patrick Schramowski, Srivatsa~R Kundurthy, Katherine Crowson, Ludwig Schmidt, Robert Kaczmarczyk, and Jenia Jitsev.
\newblock {LAION}-5b: An open large-scale dataset for training next generation image-text models, 2022.

\bibitem[Seshadri et~al.(2023)Seshadri, Singh, and Elazar]{bias_amplification}
Preethi Seshadri, Sameer Singh, and Yanai Elazar.
\newblock The bias amplification paradox in text-to-image generation.
\newblock \emph{arXiv preprint arXiv:2308.00755}, 2023.

\bibitem[Szegedy et~al.(2014)Szegedy, Liu, Jia, Sermanet, Reed, Anguelov, Erhan, Vanhoucke, and Rabinovich]{inceptionv3}
Christian Szegedy, Wei Liu, Yangqing Jia, Pierre Sermanet, Scott~E. Reed, Dragomir Anguelov, D. Erhan, Vincent Vanhoucke, and Andrew Rabinovich.
\newblock Going deeper with convolutions.
\newblock \emph{2015 IEEE Conference on Computer Vision and Pattern Recognition (CVPR)}, pages 1--9, 2014.

\bibitem[von Platen et~al.(2023)von Platen, Patil, Lozhkov, Cuenca, Lambert, Rasul, Davaadorj, Nair, Paul, Liu, Berman, Xu, and Wolf]{diffusers2023}
Patrick von Platen, Suraj Patil, Anton Lozhkov, Pedro Cuenca, Nathan Lambert, Kashif Rasul, Mishig Davaadorj, Dhruv Nair, Sayak Paul, Steven Liu, William Berman, Yiyi Xu, and Thomas Wolf.
\newblock Diffusers: State-of-the-art diffusion models.
\newblock \url{https://github.com/huggingface/diffusers}, 2023.

\end{thebibliography}
}
\newpage
\clearpage
\appendix

\onecolumn
\section{Qualitative Examples of Diversity} \label{app:qualitative_examples}

\begin{figure}[h!]
    \centering    

\includegraphics[width=\textwidth]{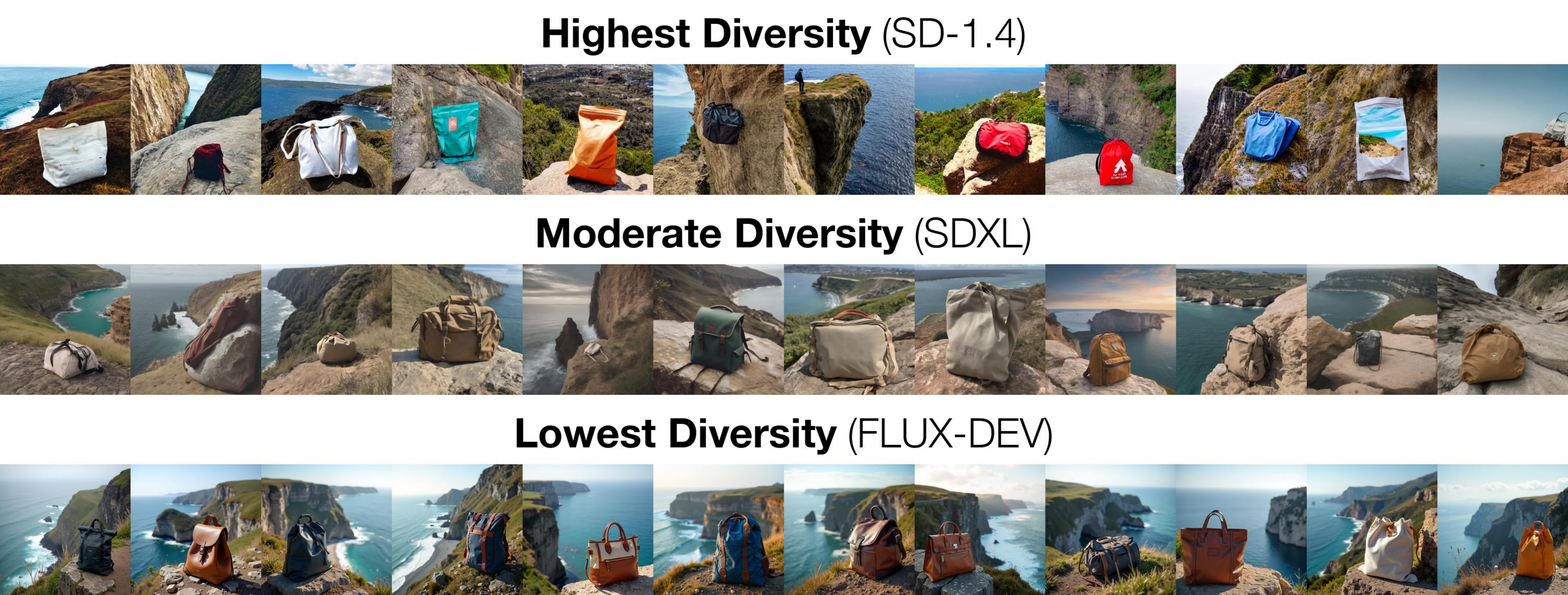}
    \caption{
    \textbf{Difference in diversity between models.} Images generated using the prompt ``a bag on a cliffside''. Each row corresponds to a model, top-down: \sdclassic{} (most diverse), \sdxl{}, and \fluxdev{} (least diverse). While no model exhibits high diversity, there is a marked difference between \sdclassic{} and \fluxdev{}, with \sdxl{} between them. Specifically, diversity is reduced in attributes such as color and placement of the bags, as well as the background.}

    \label{fig:bag_diversity_fig}
\end{figure}

\begin{figure}[h!]
    \centering    

\includegraphics[width=\textwidth]{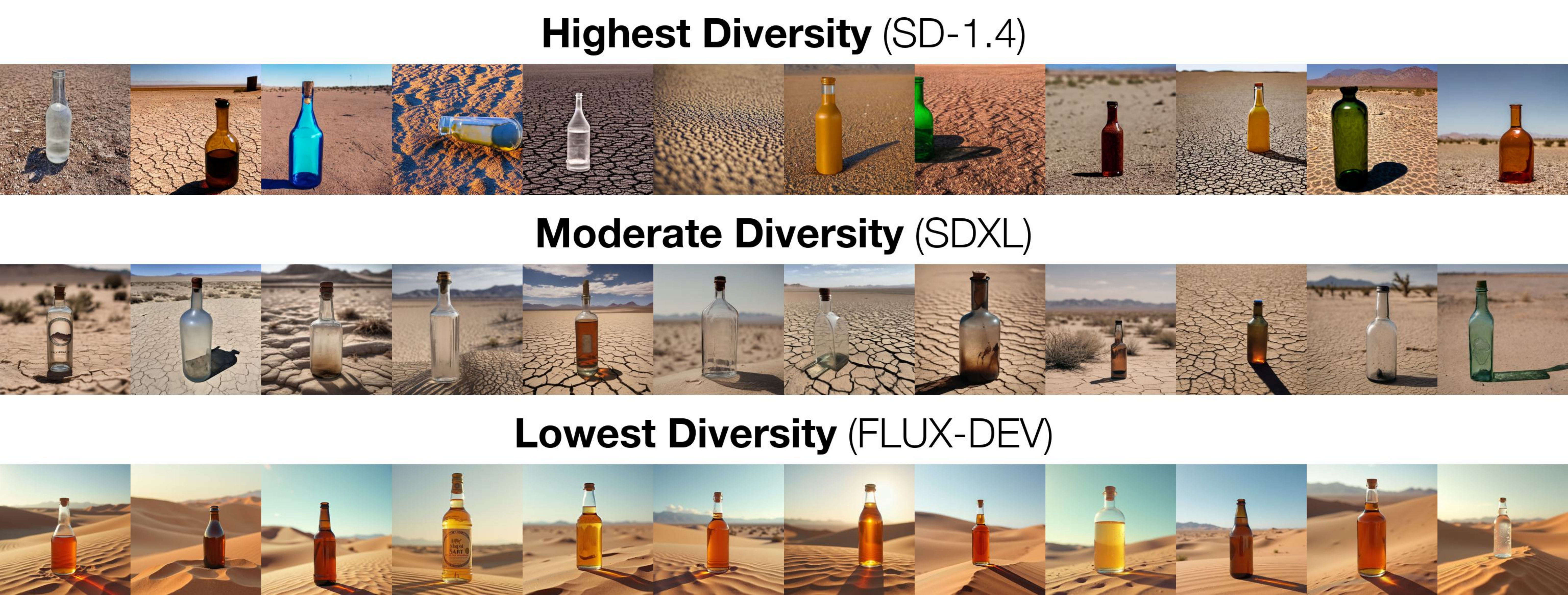}
    \caption{
    \textbf{Difference in diversity between models.} Images generated using the prompt ``a bottle in a desert''. Each row corresponds to a model, top-down: \sdclassic{} (most diverse), \sdxl{}, and \fluxdev{} (least diverse). While no model exhibits high diversity, there is a marked difference between \sdclassic{} and \fluxdev{}, with \sdxl{} between them. Here, the lack of diversity is most pronounced in the color of the bottle or its liquid. While \sdclassic{} depicts relatively varied bottles, \sdxl{} depicts transparent ones, while \fluxdev{} depicts almost exclusively orange-like bottles.}

    \label{fig:bottle_diversity_fig}
\end{figure}

\begin{figure}[h!]
\centering    
\includegraphics[width=\textwidth]{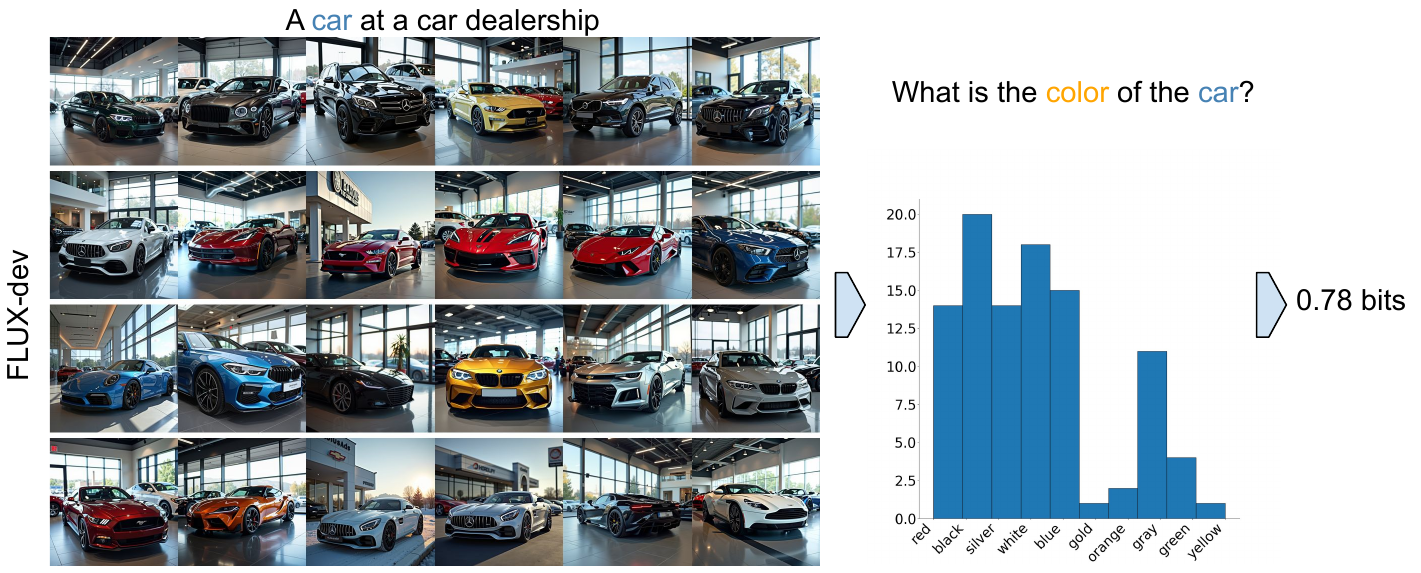}
    \caption{
    \textbf{Illustration of GRADE score.} Displayed are 24 of the \numImagesPerPrompt{} images generated by \fluxdev{} using the prompt ``A car in a car dealership''. The accompanying histogram and the subsequent entropy plot both represent the \numImagesPerPrompt{} sample. The GRADE score is 0.78, indicating the color of the cars is relatively diverse.
    }
    \label{fig:demonstrating_grade_car}
\end{figure}

\begin{figure}[h!]
\centering    
\includegraphics[width=\textwidth]{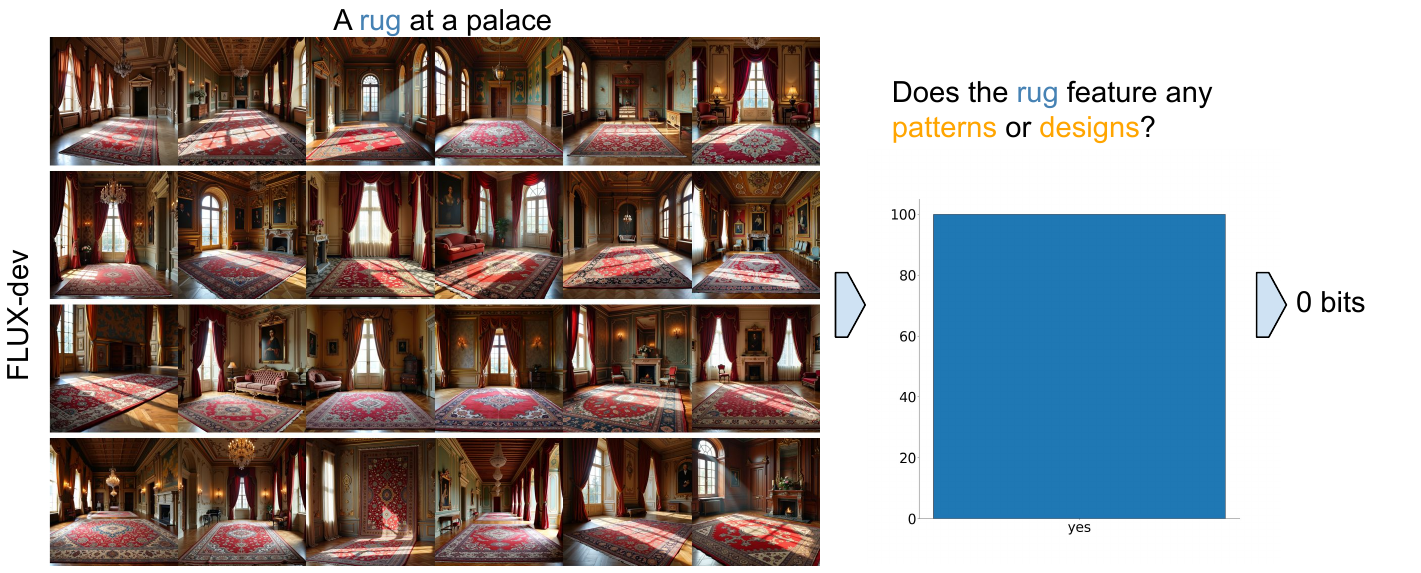}
    \caption{
    \textbf{Illustration of GRADE score.} Displayed are 24 of the \numImagesPerPrompt{} images generated by \fluxdev{} using the prompt ``A rug at a palace''. The accompanying histogram and the subsequent entropy plot both represent the \numImagesPerPrompt{} sample. The GRADE score is 0, indicating the rugs are consistently patterned.
    }
    \label{fig:demonstrating_grade_rug}
\end{figure}

\begin{figure}[h!]
\centering    
\includegraphics[width=\textwidth]{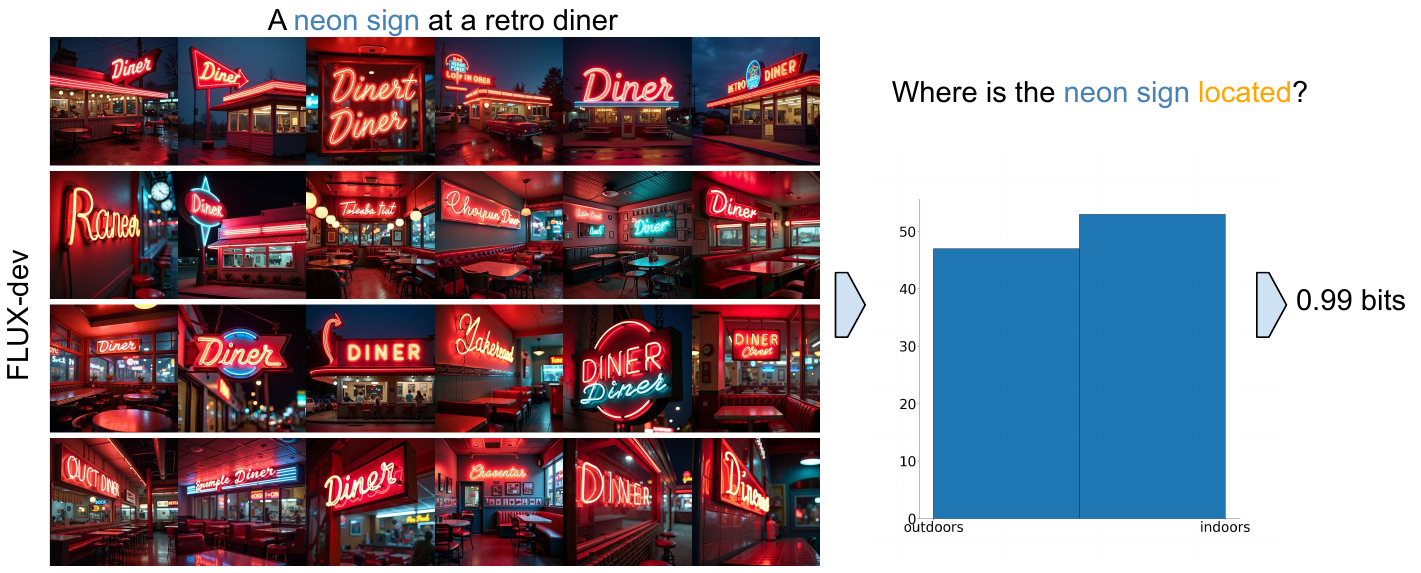}
    \caption{
    \textbf{Illustration of GRADE score.} Displayed are 24 of the \numImagesPerPrompt{} images generated by \fluxdev{} using the prompt ``A neon sign at a retro diner''. The accompanying histogram and the subsequent entropy plot both represent the \numImagesPerPrompt{} sample. The GRADE score is 0.99, indicating the location of the signs is uniform.
    }
    \label{fig:demonstrating_grade_neonsign}
\end{figure}

\clearpage
\section{Extended Data Overview} \label{app:extended_data_overview}

\begin{table}[ht]

\centering
\begin{tabular}{@{} >{\raggedright\arraybackslash}p{2cm} 
                >{\raggedright\arraybackslash}p{4cm} 
                >{\raggedright\arraybackslash}p{7cm} @{}}

\toprule

\textbf{Concept} & \textbf{Attribute} & \textbf{Attribute Values} \\ 
\midrule

\multirow{2}{2cm}{Bin} 
  & What shape is the bin? 
  & circular, octagonal, square, cylindrical, triangular, rectangular, round, oval, hexagonal \\
  & What material is the bin made from? 
  & mesh, cardboard, carbon fiber, rubber, wood, bamboo, wicker, plastic, ceramic, stainless steel, fiberglass, metal, aluminum, steel, fabric, glass \\
  & Does the bin have a lid? 
  & yes, no \\
% \hline
\multirow{2}{2cm}{Person} 
  & Is the person male or female? 
  & male, female \\
  & Does the image show the person from up-close? 
  & yes, no \\
% \hline
\multirow{2}{2cm}{Suitcase} 
  & Is the suitcase open or closed? 
  & open, closed \\
  & Is the suitcase soft-shell or hard-shell? 
  & soft-shell suitcase, hard-shell suitcase \\
% \hline
\multirow{2}{2cm}{Cake} 
  & Does the cake have multiple tiers? 
  & yes, no \\
  & Is the cake eaten? 
  & yes, no \\
  & What flavor is the cake?
  & tiramisu, cheesecake, carrot, chocolate, strawberry, vanilla \\
% \hline
\multirow{2}{2cm}{Pool} 
  & Is there anyone swimming in the pool? 
  & yes, no \\
  & What color is the water in the pool? 
  & reflective like a mirror, black, clear, green, blue, brown \\
% \hline
\multirow{2}{2cm}{Teapot} 
  & What shape is the teapot?
  & rectangular, spherical, oval, round, square, cylindrical \\
% \hline
\multirow{2}{2cm}{Bear} 
  & What species of bear is depicted in the image? 
  & polar bear, black bear, sloth bear, grizzly bear, sun bear, panda bear \\
\bottomrule
\end{tabular}
\caption{\textbf{Sample of concepts, attributes, and attribute values.} Each concept-attribute pair is a multi-prompt distribution.}
\label{tab:concept_attributes}
\end{table}

\clearpage
\section{Comparing \grade{} to Previous Metrics} \label{app:comparing_grade_to_prev_metrics}
\begin{table}[h]

  \centering
  
  \begin{tabular}{l@{\hskip 4pt}l@{\hskip 4pt}c@{\hskip 4pt}c@{\hskip 4pt}c@{\hskip 4pt}c@{\hskip 4pt}c@{\hskip 4pt}c}
    \toprule
    \textbf{Model} & \textbf{Dataset} & \textbf{\fid{}-R} & \textbf{\fid{}-P} & \textbf{R-P} & 
    \textbf{\fid{}-TVD} & \textbf{R-TVD} & \textbf{P-TVD} \\
    \midrule
    \sdclassicfirstckpt{} & \laiontwob{}   & $0.14$ & $-0.15$ & $0$ & $0.12$ & $-0.15$ & $0$ \\
    \sdclassic{}          & \laiontwob{}   & $0.19$ & $-0.40$ & $0$ & $0$ & $-0.20$ & $-0.15$ \\
    \sdtwo{}              & \laionfiveb{}  & $-0.21$ & $-0.48$ & $0$ & $0$ & $-0.19$ & $0.15$ \\
    % Add more rows as needed
    \bottomrule
  \end{tabular}
    \caption{\textbf{PCC between \grade{} and traditional metrics paired with \clip{}.} \fid{}, \recall{} (R), and \precision{} (P) show low to moderate degrees of correlation among each other, while the TVD based on the distributions from \grade{} exhibits weak correlations with all of them. This indicates the distributions estimated by \grade{} capture diversity existing metrics do not.}
  \label{tab:full_correlation_to_existing_metrics_clip}
\end{table}

\begin{table}[h]

\centering

\begin{tabular}{llrrrr}
\toprule
\textbf{Model} & \textbf{Dataset} & \textbf{TVD} & \textbf{\fid{}} & \textbf{Recall} & \textbf{Precision} \\
\midrule
\sdclassicfirstckpt{} & \laiontwob{} & $0.15$ & $290$ & $0.12$ & $0.88$ \\
\sdclassic{} & \laiontwob{} & $0.15$ & $276$ & $0.15$ & $0.92$ \\
\sdtwo{} & \laionfiveb{} & $0.16$ & $290$ & $0.12$ & $0.94$ \\
% Add more rows as needed
\bottomrule
\end{tabular}
\caption{\textbf{Evaluation results with traditional metrics paired with \clip{}.} Each value in the table is the mean of the metric over the 50 pairs of multi-prompt distributions.}
\label{tab:evaluation_compare_existing_metrics_clip}
\end{table}

% \begin{table}[h]

% \centering

% \begin{tabular}{l@{\hskip 4pt}l@{\hskip 4pt}c@{\hskip 4pt}c@{\hskip 4pt}c@{\hskip 4pt}c@{\hskip 4pt}c@{\hskip 4pt}c} % Changed: used c for center-alignment, added space reduction between columns
% \toprule
% \textbf{Model} & \textbf{Dataset} & \textbf{\fid{}-R} & \textbf{\fid{}-P} & \textbf{R-P} & \textbf{\fid{}-TVD} & \textbf{R-TVD} & \textbf{P-TVD} \\
% \midrule
% \sdclassicfirstckpt{} & \laiontwob{}   & -0.41 & 0.23 & -0.34 & 0.14 & 0.04 & 0 \\
% \sdclassic{}          & \laiontwob{}   & -0.48 & 0.14 & -0.22 & 0.18 & -0.10 & 0.14 \\
% \sdtwo{}              & \laionfiveb{}  & -0.12 & -0.52 & 0 & -0.16 & -0.15 & 0.13 \\

% \bottomrule
% \end{tabular}
% \caption{\textbf{PCC between \grade{} and traditional metrics paired with \inception{}.} \fid{}, \recall{} (R), and \precision{} (P) show low to moderate degrees of correlation among each other, while the TVD based on the distributions from \grade{} exhibits weak correlations with all of them. This indicates the distributions estimated by \grade{} capture diversity existing metrics do not.}
% \label{tab:correlation_to_existing_metrics_inception}
% \end{table}

\begin{table}[h]

\centering

\begin{tabular}{l@{\hskip 4pt}l@{\hskip 4pt}c@{\hskip 4pt}c@{\hskip 4pt}c@{\hskip 4pt}c@{\hskip 4pt}c@{\hskip 4pt}c} % Changed: used c for center-alignment, added space reduction between columns
\toprule
\textbf{Model} & \textbf{Dataset} & \textbf{\fid{}-R} & \textbf{\fid{}-P} & \textbf{R-P} & \textbf{\fid{}-TVD} & \textbf{R-TVD} & \textbf{P-TVD} \\
\midrule
\sdclassicfirstckpt{} & \laiontwob{}   & $-0.41$ & $0.23$ & $-0.34$ & $0.14$ & $0.04$ & $0$ \\
\sdclassic{}          & \laiontwob{}   & $-0.48$ & $0.14$ & $-0.22$ & $0.18$ & $-0.10$ & $0.14$ \\
\sdtwo{}              & \laionfiveb{}  & $-0.12$ & $-0.52$ & $0$ & $-0.16$ & $-0.15$ & $0.13$ \\

\bottomrule
\end{tabular}
\caption{\textbf{PCC between \grade{} and traditional metrics paired with \inception{}.} \fid{}, \recall{} (R), and \precision{} (P) show low to moderate degrees of correlation among each other, while the TVD based on the distributions from \grade{} exhibits weak correlations with all of them. This indicates the distributions estimated by \grade{} capture diversity existing metrics do not.}
\label{tab:correlation_to_existing_metrics_inception}
\end{table}

\begin{table}[h]

\centering

\begin{tabular}{llrrrr}
\toprule
\textbf{Model} & \textbf{Dataset} & \textbf{\tvdgrade{}} & \textbf{\fid{}} & \textbf{Recall} & \textbf{Precision} \\
\midrule
\sdclassicfirstckpt{} & \laiontwob{} & $0.15$ & $19.67$ & $0.35$ & $0.75$ \\
\sdclassic{} & \laiontwob{} & $0.15$ & $15.0$ & $0.45$ & $0.74$ \\
\sdtwo{} & \laionfiveb{} & $0.16$ & $18.67$ & $0.49$ & $0.83$ \\
% Add more rows as needed
\bottomrule
\end{tabular}
\caption{\textbf{Evaluation results with existing metrics using \inception{}.} Each value in the table is the mean of the metric over the 50 pairs of distributions.}
\label{tab:evaluation_compare_existing_metrics_inception}
\end{table}

Extended results for the comparison between \grade{} and traditional metrics described in \cref{subsec:correlation_to_existing_metrics}. Results using \clip{} for feature extraction can be viewed in \cref{tab:full_correlation_to_existing_metrics_clip} and \cref{tab:evaluation_compare_existing_metrics_clip}. Results using \inception{} \citep{inceptionv3} (ImageNet features \citep{imagenet}) are in \cref{tab:correlation_to_existing_metrics_inception} and \cref{tab:evaluation_compare_existing_metrics_inception}. Below we detail the process of collecting the image sets and comparing between them.

\paragraph{Reference and generated images.}
Since LAION is opensource and was used to train \sdclassicfirstckpt{}, \sdclassic{}, and \sdtwo{}; \laiontwob{} for the first two and \laionfiveb{} for the latter--we sample images from it and compare them to images generated by the models. Specifically, we sample 50 of the \numConceptDistributions{} multi-prompt distributions (that is, only the concept, attribute, and attribute values, not the prompts and images) in \cref{sec:comparing_t2i_models}. Next, we sample 115 image and caption pairs using \wimbd{}, where the image depicts the concept and the caption mentions the concept but not the attribute, in accordance with our approach (\cref{sec:problem_description}). We end up with 50 reference distributions, each consisting of 115 images. To get our generated images, we generate one image for each caption, to maintain equal proportion between the distributions. For example, if an image in LAION is linked to the caption ``Unicorn Cookie'', its corresponding distribution will contain an image that was generated using that caption as a prompt.

\paragraph{Details of metrics.} Using the 50 pairs of distributions, we can compare \grade{} to the metrics. Since entropy is not a reference-based metric, we change it in favor of Total Variation Distance (TVD) and use it on top of the distributions estimated by \grade{}. We compute \fid{} and \recall{}, using features from the open-clip implementation (the \texttt{ViT-H/14} variant) \citep{openai_clip_software, clipscore}, trained on \laiontwob{}. \recall{} was computed with $k=3$. We run the same experiment using \inception{} features with $64$ dimensions.

\subsection{Qualitative Metric Comparison Examples} \label{app:qualitative_metric_comparison_examples}

\begin{figure}[h]
    \centering
    \includegraphics[width=1.0\textwidth]{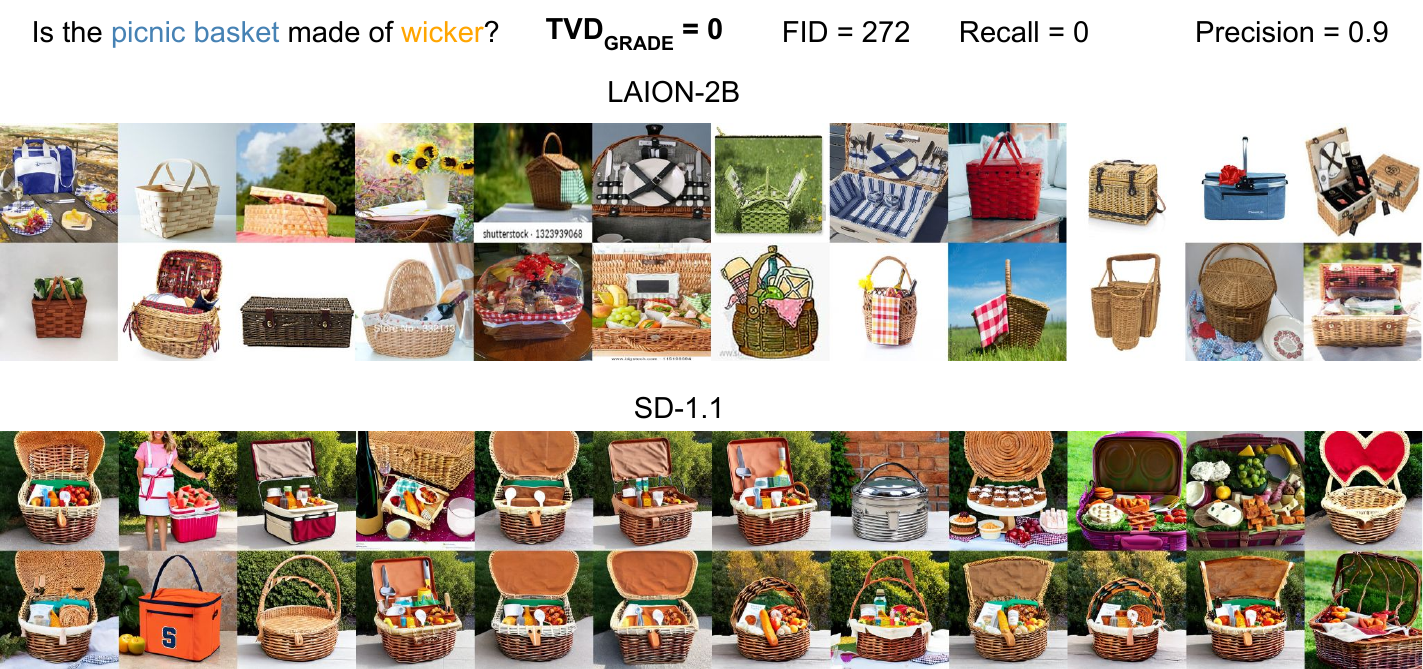}
    \caption{\textbf{Comparison between \grade{}, \fid{}, and \recall{}, using \clip{} features.} The metrics are compared over the ``wicker'' attribute of the concept ``picnic basket''. TVD\textsubscript{GRADE} reports very high similarity (lower TVD is better) between the sets of images, which is indeed shown in the images (almost all picnic baskets are made of wicker). In contrast, \recall{} and \fid{} report very low scores.}
\label{fig:compare_fid_recall_to_grade_sd11}
\end{figure}

\begin{figure}[h]
    \centering
    \includegraphics[width=1.0\textwidth]{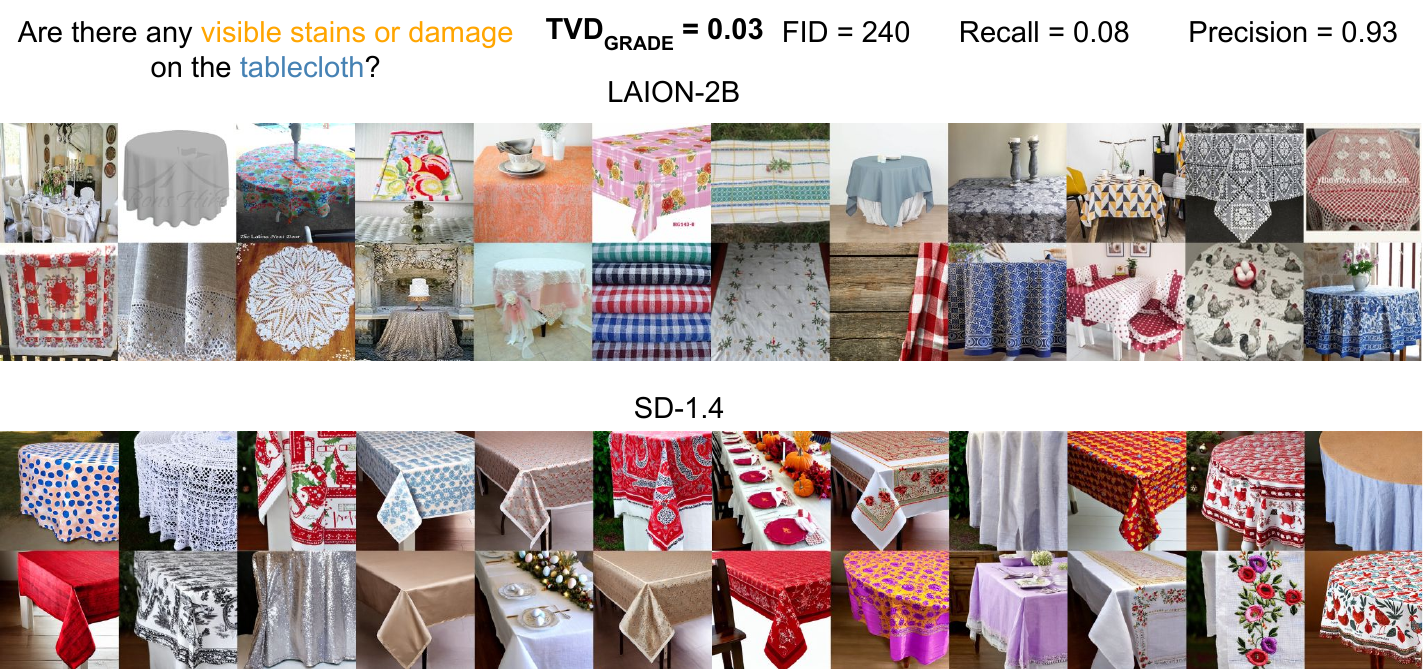}
    \caption{\textbf{Comparison between \grade{}, \fid{}, and \recall{}, using \clip{} features.} The metrics are compared over the ``visible stains or damage'' attribute of the ``tablecloth'' concept. TVD\textsubscript{GRADE} reports very high similarity (lower TVD is better) between the sets of images, which is indeed shown in the images (the tablecloth is rarely damaged in either set). In contrast, \recall{} and \fid{} report very low scores.}
\label{fig:compare_fid_recall_to_grade_sd14}
\end{figure}

\clearpage
\section{Extended Diversity Comparisons between T2I Models} \label{app:extended_results}

\begin{figure}[h]
    \centering
    % First Subfigure
    \begin{subfigure}[b]{0.48\textwidth}
        \centering
        \includegraphics[width=\textwidth]{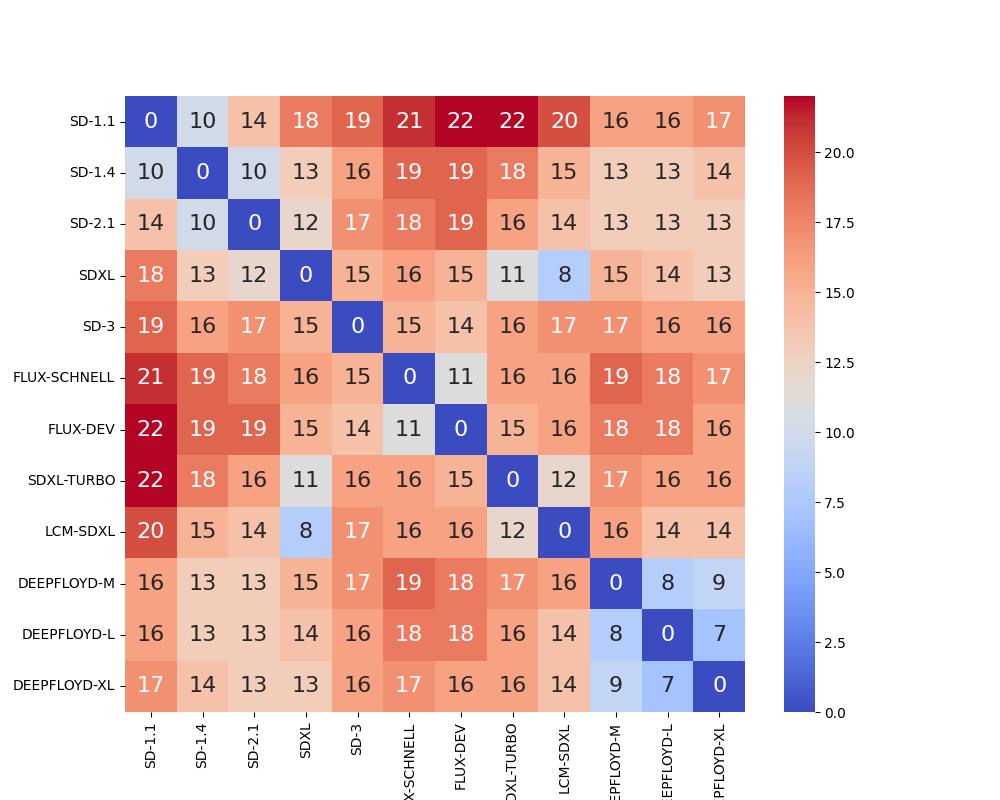}
        % \caption{\textbf{The mean TVD between all pairs of models over multi-prompt distributions.}}
    \end{subfigure}
    \hfill
    % Second Subfigure
    \begin{subfigure}[b]{0.48\textwidth}
        \centering
        \includegraphics[width=\textwidth]{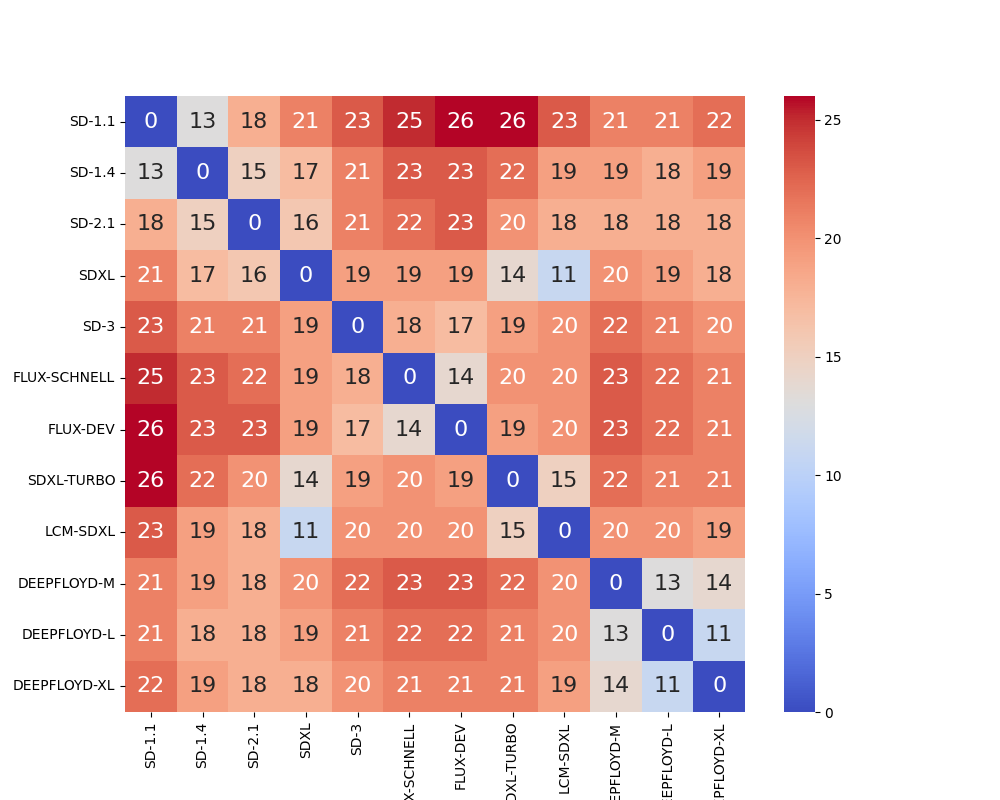}
        % \caption{\textbf{The mean TVD between all pairs of models over single prompt distributions.}}
    \end{subfigure}
    \caption{\textbf{The mean total variation distance (TVD) between all pairs of models over (a) multi-prompt distributions and (b) single prompt distributions.} For readability, both figures show TVD in a range between 0 and 100 instead of 0 to 1.}
    \label{fig:combined_app_tvd}
\end{figure}

\begin{table}[h]
    \centering
    
    \begin{tabular}{llr}
        \toprule
        \textbf{Backbone} & \textbf{Models} & \textbf{Mean TVD}\\
        \midrule
        \sdclassicfirstckpt{}      & \sdclassicfirstckpt{}, \sdclassic{}, \sdtwo{} & $11$\\
        \sdxl{}    & \sdxl{}, \sdxllcm{}, \sdxl{} & $10$ \\
        FLUX    & \fluxschnell{}, \fluxdev{} & $11$ \\
        DeepFloyd & \deepfloydM{}, \deepfloydL{}, \deepfloydXL{} & $8$\\
        \bottomrule
    \end{tabular}

    \caption{\textbf{Backbones, their associated models, and the mean TVD of models with a shared backbone.}}
    \label{tab:backbone_models_app}
\end{table}

\textbf{Similarity in diversity across distributions.} We investigate the similarity in diversity across models we find in \cref{subsec:results}. We modify \grade{} to use Total Variation Distance (TVD) instead of entropy to facilitate comparisons between corresponding distributions in the attribute value level. For example, the difference between the frequency of ``blue'' in the multi-prompt distribution of the concept \emph{tie} and attribute \emph{color}. Results for both multi and single prompt distributions are shown in \cref{fig:combined_app_tvd}. The results are in line with our other findings: all models have similar distributions, with the maximum TVD for multi-prompt distributions being 0.22 and for single prompt distributions 0.26, with these numbers being the result of a comparison between the least and most diverse models (i.e., \sdclassicfirstckpt{} and \fluxdev{}). Moreover, models with similar backbone have smaller TVDs. The groups and the mean TVDs are shown in \cref{tab:backbone_models_app}.

\subsection{Additional Analysis on Model Size}
\begin{figure}[h]
    \centering
    \begin{subfigure}[t]{0.48\textwidth}
        \centering
        % \adjustbox{max width=\linewidth,max height=5cm}{\includegraphics{Figures/concept_inverse_scale_law.pdf}}

        \adjustbox{max width=\linewidth,max height=5cm}{\includegraphics{Figures/scale_law_plot_prompt_level_False.pdf}}

        \label{fig:app_concept_inverse_scale_law_fig}
        % \caption{Model diversity score in multi-prompt setting plotted against the denoiser's parameter size.}
    \end{subfigure}
    \hfill
    % Second Subfigure
    \begin{subfigure}[t]{0.48\textwidth}
        \centering
        % \adjustbox{max width=\linewidth,max height=5cm}{
        %     \includegraphics[width=\textwidth]{Figures/prompt_inverse_scale_law.pdf}
        % }

        \adjustbox{max width=\linewidth,max height=5cm}{
            \includegraphics[width=\textwidth]{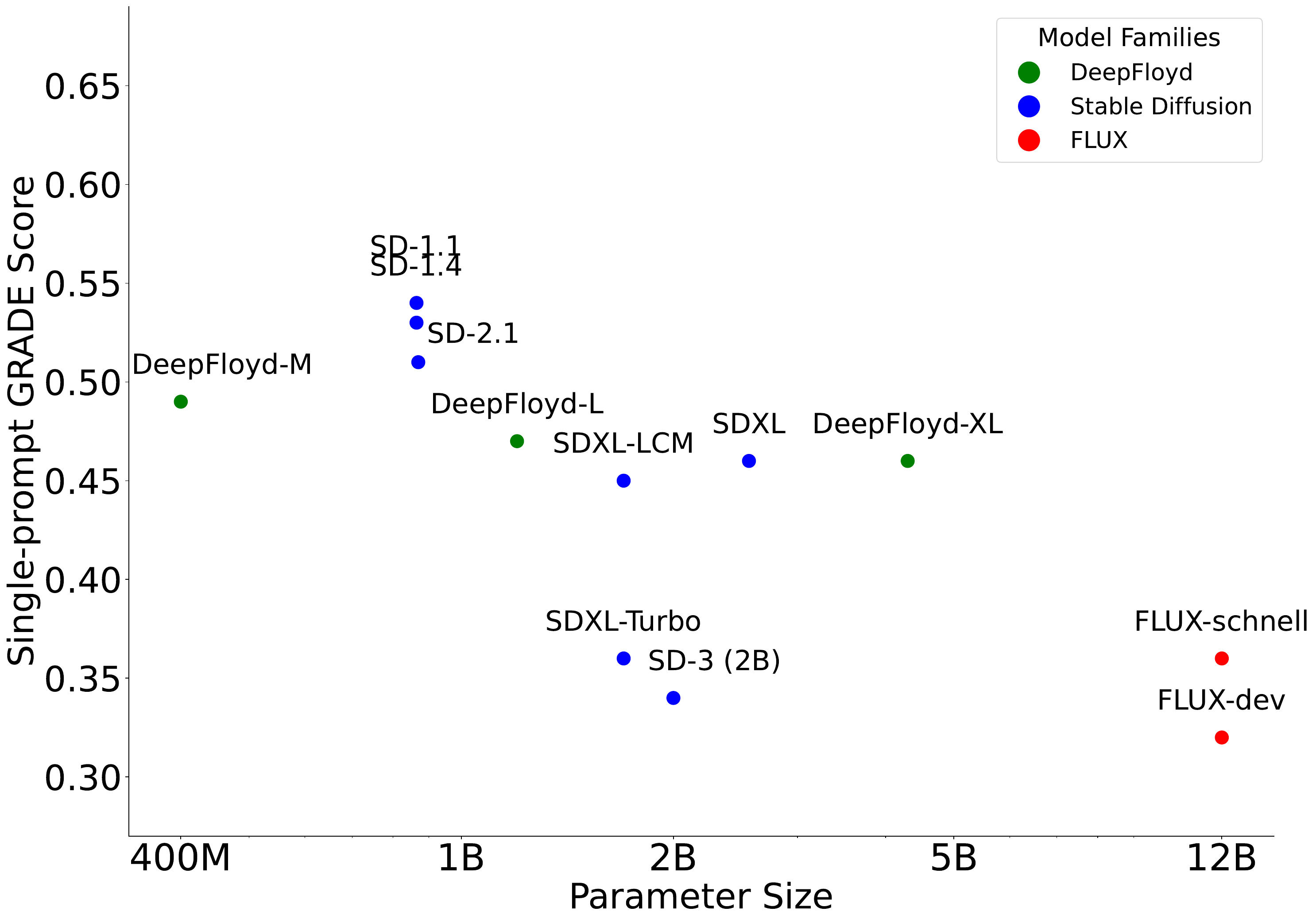}
        }
        \label{fig:app_prompt_inverse_scale_law_fig}
        % \caption{Model diversity score in single-prompt setting plotted against the denoiser's parameter size.} 
    \end{subfigure}
    \caption{\textbf{(a) GRADE score in multi-prompt setting plotted against the denoiser's parameter size. (b) GRADE score in single-prompt setting plotted against the denoiser's parameter size.} To a degree, diversity deteriorates in tandem with parameter size. This phenomenon is most apparent within every model family.}
    \label{fig:combined_app_inverse_scale_law}
\end{figure}

\begin{figure}[h]
    \centering
    % First Subfigure
    \begin{subfigure}[b]{0.48\textwidth}
        \centering
        % \adjustbox{max width=\linewidth,max height=5cm}{\includegraphics{Figures/concept_level_diversity_vs_pa.pdf}} 
        \adjustbox{max width=\linewidth,max height=5cm}{\includegraphics{Figures/NEW_prompt_adherence_vs_diversity_plot_promptlevel_False.pdf}} 
        % \caption{Model diversity in multi-prompt setting plotted against the \% of ``none of the above''.}
    \end{subfigure}
    \hfill
    % Second Subfigure
    \begin{subfigure}[b]{0.48\textwidth}
        \centering
        % \adjustbox{max width=\linewidth,max height=5cm}{\includegraphics{Figures/prompt_level_diversity_vs_pa.pdf}}   
        \adjustbox{max width=\linewidth,max height=5cm}{\includegraphics{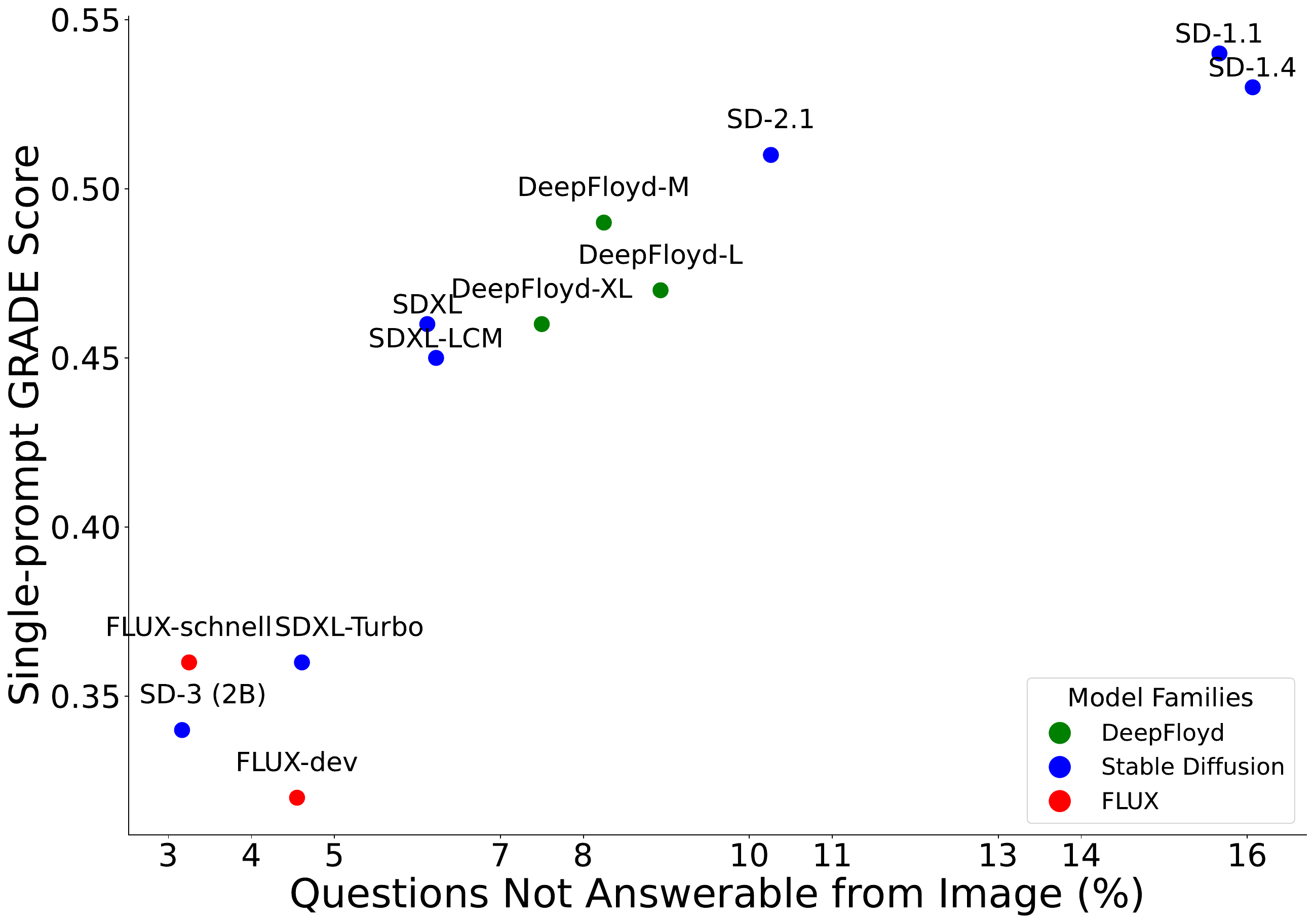}}   
        
        % \caption{Model diversity in single-prompt setting plotted against the \% of ``none of the above''.}
    \end{subfigure}
    \caption{\textbf{(a) GRADE score in multi-prompt setting plotted against the \% of ``none of the above''. (b) GRADE score in single-prompt setting plotted against the \% of ``none of the above''.} In \cref{subsec:validating_grade} we show 80\% of which account for missing concepts in the image. The plots show negative correlation between diversity and prompt adherence, which indicates there is a tradeoff.}
    \label{fig:combined_app_pa_vs_entropy}
\end{figure}

We further investigate the relationship between model size and diversity, and prompt adherence and diversity. \cref{fig:combined_app_inverse_scale_law} shows that as the denoisers' parameter size increases, the GRADE scores in both the multi and single prompt distributions decrease. This suggests that larger models produce less diverse outputs, indicating an inverse-scaling law \citep{inverse_scale_law}. The negative correlation is supported by significant Pearson and Spearman correlation coefficients at both the concept level (Pearson $r = -0.701$, $p = 0.011$; Spearman $\rho = -0.842$, $p = 0.001$) and the prompt level (Pearson $r = -0.666$, $p = 0.018$; Spearman $\rho = -0.804$, $p = 0.002$).

\Cref{fig:combined_app_pa_vs_entropy} illustrates negative correlation between diversity and prompt adherence. As the percentage of unanswerable images (``none of the above'') increases i.e., prompt adherence \emph{decreases}, the diversity measured by entropy increases. This is quantified by strong positive Pearson and Spearman correlations at both the concept level (Pearson $r = 0.802$, $p = 0.002$; Spearman $\rho = 0.938$, ($p < 0.001$) and the prompt level (Pearson $r = 0.871$, ($p < 0.001$); Spearman $\rho = 0.947$, ($p < 0.001$). This indicates a trade-off between diversity and prompt adherence: models that generate more diverse outputs tend to adhere less strictly to the prompts.

\subsection{Statistical Significance of GRADE scores}
\label{app:subsec_statistical_sig_exp}

To confirm our results are statistically significant, we perform a two-tailed permutation test between every unique pair of models for both distribution types (single-prompt and multi-prompt). This test is common when the data comes from a complex distribution \citep{bonnini2024review}, in our case, the distribution of GRADE scores of each model. We demonstrate that the difference between the vast majority of models is statistically significant in both cases.

Concretely, there are 66 unique model pairs. For each pair, we compute a two-tailed permutation test with the null hypothesis \( H_0 \) that the GRADE scores of the two models are the same. We perform \( N = 100,\!000 \) permutations, where the p-value is defined as:
\[
p = \frac{\text{number of permutations where } |D_{\text{perm}}| \geq |D_{\text{obs}}|}{N},
\]
where \( D_{\text{obs}} \) is the observed difference in GRADE scores between the two models, and \( D_{\text{perm}} \) is the difference obtained under each permutation. We compare the p-value \( p \) to a significance level of \( \alpha = 0.05 \).

\paragraph{Results.} The vast majority of pairs are statistically significant.

Comparisons based on single-prompt distributions reveal just three pairs are not statistically significant:\\ (\sdxl{}, \sdxllcm{}), (\sdxl{}, \deepfloydXL{}), and (\sdxlturbo{}, \fluxschnell{}). 

Similarly, comparisons using multi-prompt distributions, reveal only 15 pairs are not statistically significant:\\ (\sdclassicfirstckpt{}, \sdclassic{}), (\sdclassicfirstckpt{}, \sdtwo{}), (\sdclassicfirstckpt{}, \deepfloydM{}), (\sdclassicfirstckpt{}, \deepfloydL{}), (\sdclassic{}, \sdtwo{}), (\sdclassic{}, \deepfloydM{}), (\sdclassic{}, \deepfloydL{}),  (\sdtwo{}, \deepfloydM{}), (\sdtwo{}, \deepfloydL{}), (\sdtwo{}, \deepfloydXL{}), (\sdxl{}, \sdxllcm{}), (\deepfloydL{}, \deepfloydXL{}), (\sdvit{}, \fluxschnell{}), (\sdvit{}, \fluxdev{}), and (\fluxschnell{}, \fluxdev{}).

Non-significant pairs are similar in quality. For example, all pair combinations of SD-1.1, SD-1.4, and SD-2.1 are not significant, which is not surprising since these models largely share the same underlying architectures and training data.

\begin{figure}[t!]
    \centering
    \begin{subfigure}[t]{0.48\textwidth}
        \centering
        \adjustbox{max width=\linewidth,max height=5cm}{\includegraphics{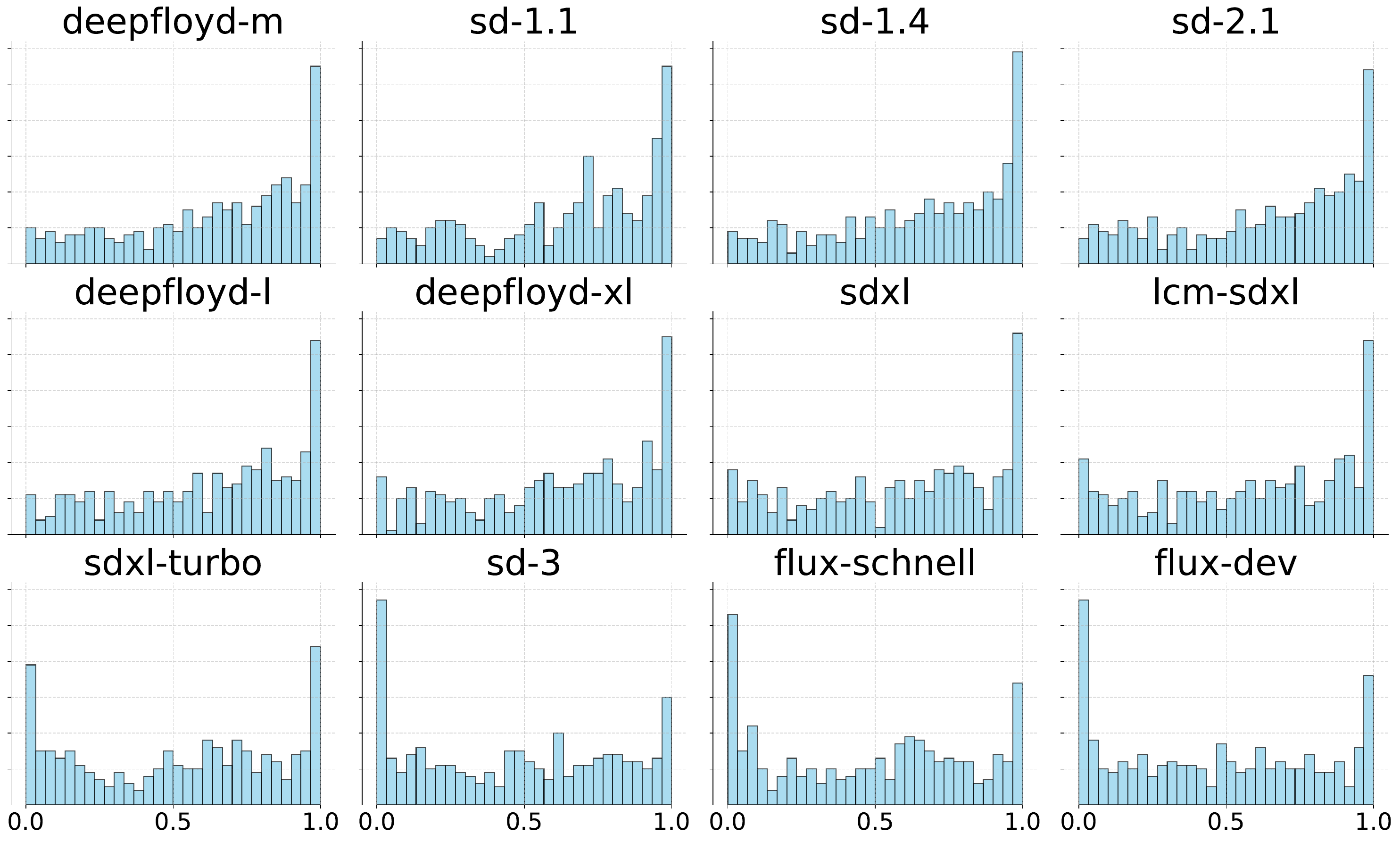}}
        \label{fig:concept_entropy_histograms}
        \caption{\textbf{Histogram from each model in the multi-prompt setting.}}
    \end{subfigure}
    \hfill
    % Second Subfigure
    \begin{subfigure}[t]{0.48\textwidth}
        \centering
        \adjustbox{max width=\linewidth,max height=5cm}{
            \includegraphics[width=\textwidth]{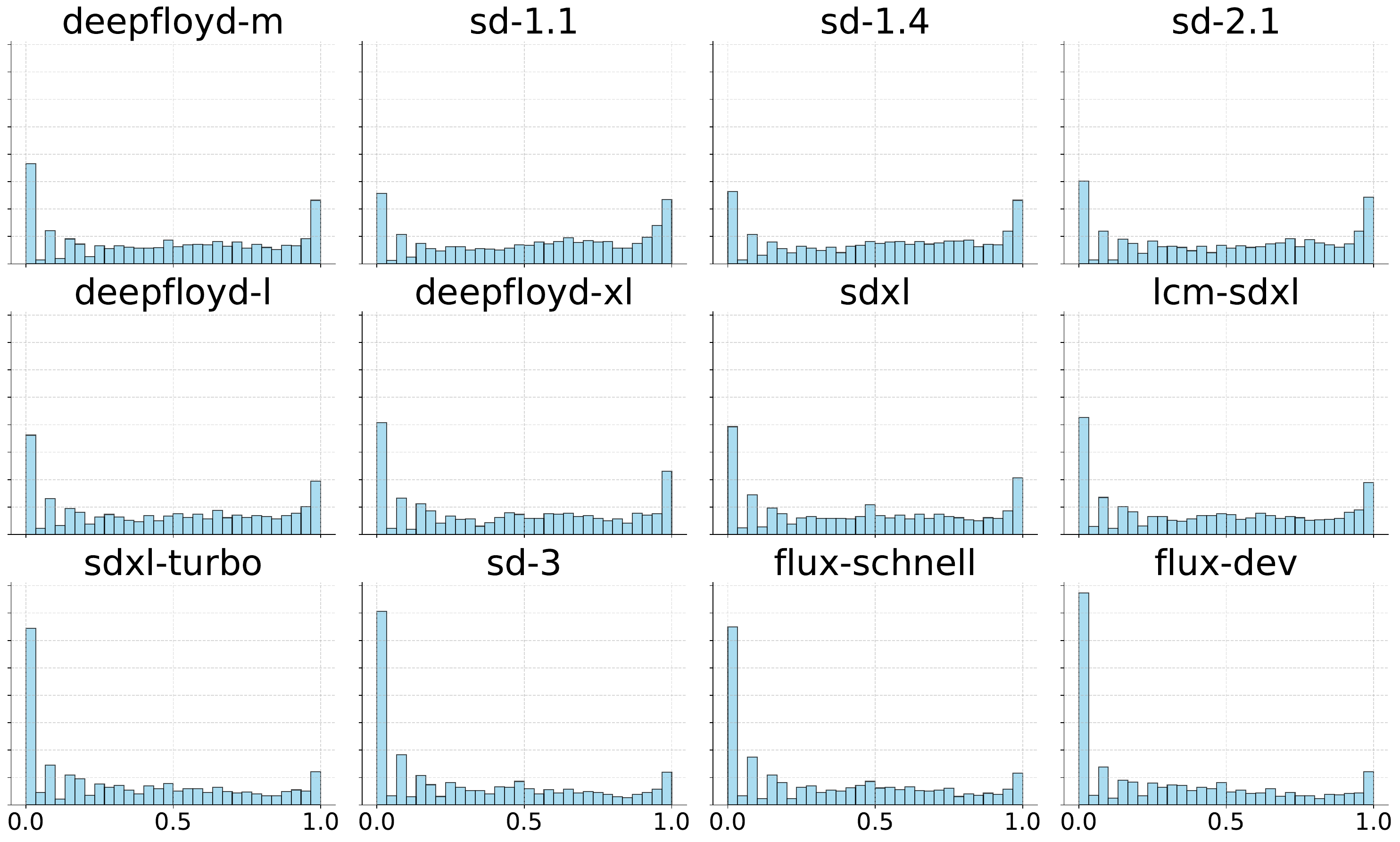}
        }
        \label{fig:prompt_entropy_histograms}
        \caption{\textbf{Histogram from each model in the single-prompt setting.}} 
    \end{subfigure}
    \caption{\textbf{A histogram of the GRADE scores (normalized entropy) from each model for both distribution types.} Except the histograms of the most diverse models in the multi-prompt setting, histograms exhibit bimodal distributions, with peaks near both tails.}
    \label{fig:mean_entropy_histograms}
\end{figure}

 \paragraph{Why standard deviation can be misleading.} We report standard deviations for completeness in \cref{tab:entropy_of_models_with_std}, but emphasize that they can be uninformative for multi‐modal or heavily skewed distributions. A single standard deviation hides whether most values cluster around a single region or split between two (or more) distinct clusters, producing a deceptively large overall variance. Indeed, in \cref{fig:mean_entropy_histograms}, many models show a pronounced high‐low split in their GRADE scores. This structure is lost if one relies solely on a single summary statistic, so we encourage readers to consult the histograms.

\begin{table}[h]

\centering
% \resizebox{\columnwidth}{!}{
\begin{tabular}{@{}lcc@{}}
\toprule
& \multicolumn{2}{c}{\textbf{GRADE Score $\uparrow$}} \\
\cmidrule(lr){2-3}
\textbf{Model} & \textbf{Multi-prompt} & \textbf{Single-prompt} \\
\midrule
\deepfloydM{} & \( \mathbf{0.64 \pm 0.30}\) &  \( 0.49 \pm 0.34\)\\
\deepfloydL{} & \( 0.62 \pm 0.29\) & \(0.47 \pm 0.34\) \\
\deepfloydXL{} & \(0.61 \pm 0.30\) & \(0.46 \pm 0.34\)\\
\sdclassicfirstckpt{} &  \( \mathbf{0.64 \pm 0.30}\) & \( \mathbf{0.54 \pm 0.33}\)\\
\sdclassic{} & \( \mathbf{0.64 \pm 0.29}\) & \( 0.53 \pm 0.33\) \\
\sdtwo{} & \(0.63 \pm 0.30\)& \(0.51 \pm 0.34\) \\
\sdxl{} & \(0.59 \pm 0.31\) & \(0.46 \pm 0.34\)\\
\sdxlturbo{} & \(0.52 \pm 0.33\)&  \(0.36 \pm 0.33\)\\
\sdxllcm{} & \(0.58 \pm 0.32\) & \(0.45 \pm 0.34\) \\
\sdvit{} & \(0.47 \pm 0.33\) & \(0.34 \pm 0.33\) \\
\fluxschnell{} & \(0.48 \pm 0.33\)& \(0.36 \pm 0.33\)\\
\fluxdev{} & \(0.47 \pm 0.33\)& \(0.32 \pm 0.32\)\\
\bottomrule
\end{tabular}

\caption{\textbf{GRADE score in multi- and single-prompt distributions.} The mean entropy over all distributions for each model over multi-prompt and single-prompt settings. All models have a standard error of $\hat{\sigma} < 0.02$ and $\hat{\sigma} < 0.001$ respectively. Values close to 1 indicate highly diverse behavior (uniform) while values close to 0 indicate highly repetitive generations. The \emph{most} diverse models are in bold.}

\label{tab:entropy_of_models_with_std}

\end{table}

\subsection{Discussion of results} 
Our findings reinforce the observations made in the main text regarding the interplay between model scale, diversity, and prompt adherence:

\textbf{Inverse-scaling law.} There is a negative correlation between diversity and model size, suggesting that increasing model parameters leads to decreased diversity. This phenomenon is most apparent within each model family and aligns with the concept of an inverse-scaling law.

\textbf{Fidelity-diversity trade-off.} The negative correlation between diversity and prompt adherence indicates a trade-off between a model's ability to generate images that match the prompt and the diversity of its outputs. This is consistent with previous findings on fidelity-diversity trade-offs \citep{classifier_guidance, improved_precision_and_recall}, where improving a model's prompt-adherence reduces the overall diversity of its outputs.

\section{Default Behaviors} \label{app:default_behavior_extended_results}

In \cref{subsec:results} we define default behaviors and mention that almost all concepts are associated with at least one default behavior, as shown in \cref{tab:default_behavior_at_least_one}. In \cref{tab:default_behavior_freq_in_total}, we report the total number of default behaviors for both types of distributions.

\begin{figure}[h]
    \centering
    \includegraphics[width=1.0\textwidth]{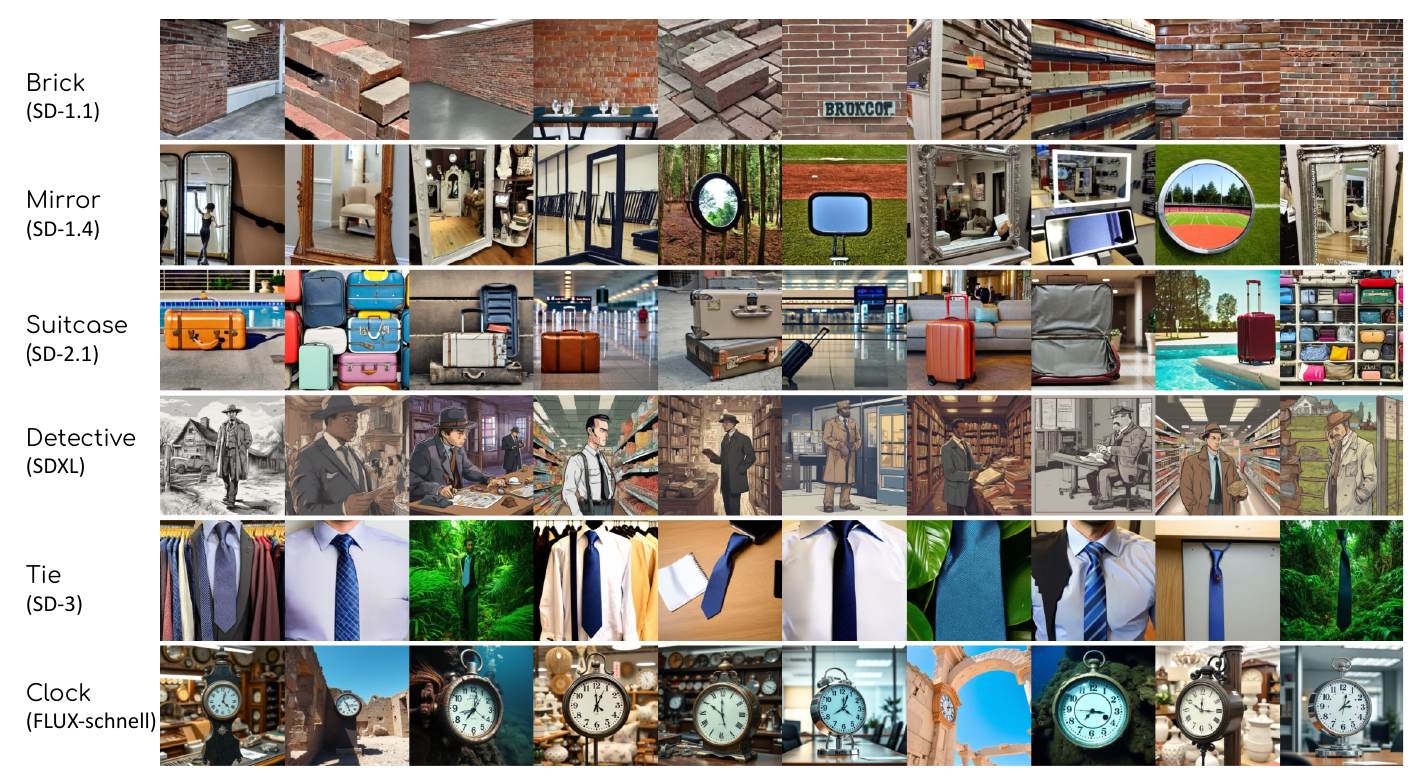}
    \caption{\textbf{A sample of images depicting the default behaviors in \cref{tab:default_behavior_samples}.} The concept is shown in the left column with the model directly below it. Images were sampled randomly from all prompts. The default behaviors, top down: (1) stacked bricks; (2) framed mirrors; (3) hard-shell suitcase; (3) male detective; (4) neckties; and (5) analog clocks.}
\label{fig:default_behavior_samples}
\end{figure}

\cref{tab:default_behavior_samples} shows a sample of default behaviors detected in multi-prompt distributions and \cref{fig:default_behavior_samples} images of these behaviors.

% Default behavior tables
\begin{table}[h]

\begin{center}
\begin{tabular}{lcc} % Centered all columns
\toprule
\multicolumn{1}{c}{} & \multicolumn{2}{c}{\bf \% of Default Behavior $\downarrow$} \\
\cmidrule(lr){2-3} % Added subtle line
\multicolumn{1}{l}{\bf Model} & \multicolumn{1}{c}{\bf Multi-prompt} & \multicolumn{1}{c}{\bf Single-prompt} 
\\ \midrule
\deepfloydM{} & $83$ & $92$ \\
\deepfloydL{} & $81$ & $92$ \\
\deepfloydXL{} & $80$ & $92$ \\
\sdclassicfirstckpt{}         & $78$ & $87$ \\
\sdclassic{}         & $82$ & $87$ \\
\sdtwo{}             & $76$ & $89$ \\
\sdxl{}              & $81$ & $90$ \\
\sdxlturbo{} & $86$ & $95$ \\
\sdxllcm{} & $82$ &  $92$ \\
\sdvit{}             & $88$ & $95$ \\
\fluxschnell{}       & $\mathbf{90}$ & $\mathbf{97}$ \\
\fluxdev{}       & $88$ & $96$ \\
\bottomrule
\end{tabular}
\end{center}
\caption{\textbf{Percentage of at least one default behavior.} Lower values indicate higher diversity. Almost all concepts are associated with at least one default behavior in single prompt distributions, with a similar trend in multi-prompt distributions. The model with the \emph{most} default behaviors is in bold. Results are rounded to the closest integer.}
\label{tab:default_behavior_at_least_one}
\end{table}

\begin{table}[h]
\begin{center}
\begin{tabular}{ccc}
\toprule
\multicolumn{1}{c}{\bf Model} & \multicolumn{2}{c}{\bf \% of Default Behavior $\downarrow$} \\
\cmidrule(lr){2-3} % Added subtle line
\multicolumn{1}{c}{} & \multicolumn{1}{c}{\bf Multi-prompt} & \multicolumn{1}{c}{\bf Single-prompt} 
\\ \midrule
\deepfloydM{} & $39$ &  $54$ \\
\deepfloydL{} & $39$ & $56$ \\
\deepfloydXL{} & $40$ & $56$ \\
\sdclassicfirstckpt{}         & $39$ & $49$ \\
\sdclassic{}         & $40$ & $51$ \\
\sdtwo{}             & $40$ & $52$ \\
\sdxl{}              & $44$ & $57$ \\
\sdxlturbo{} & $50$ &  $67$\\
\sdxllcm{} & $44$ & $57$ \\
\sdvit{}             & $56$ & $69$ \\
\fluxschnell{}       & $55$ & $67$ \\
\fluxdev{}       & $\mathbf{56}$ & $\mathbf{70}$ \\
\bottomrule
\end{tabular}
\end{center}

\caption{\textbf{Percentage of all default behaviors.} Lower values indicate higher diversity. There are 405 multi-prompt and 2430 single prompt distributions in total. The table quantifies the total percentage of default behaviors observed. The model with the \emph{most} default behaviors is in bold. Results are rounded to the closest integer.}
\label{tab:default_behavior_freq_in_total}

\end{table}

\begin{table}[t]

\begin{center}
\begin{tabular}{llll} % Four centered columns
\toprule
\textbf{Model} & \textbf{Question (Attribute)} & \textbf{Attribute Value} & \textbf{Percentage} \\
\midrule
\sdclassicfirstckpt{} & Is the \underline{brick} alone or in a stack with others? & stacked & $97.4$ \\
\sdclassic{}          & Is there a frame around the \underline{mirror}? & yes & $92.9$ \\
\sdtwo{}              & Is the \underline{suitcase} soft-shell or hard-shell? & hard-shell & $88.3$ \\
\sdxl{}               & Is the \underline{detective} female or male? & male & $99.6$ \\
\sdvit{}              & Is the \underline{tie} a necktie or a bowtie? & necktie & $100$ \\
\fluxschnell{}        & Is the \underline{clock} analog or digital? & analog & $100$ \\
\bottomrule
\end{tabular}
\end{center}

\caption{\textbf{A random sample of default behaviors.} The concept is \underline{underlined} in the question column. Images corresponding to the behaviors in the table can be viewed in \cref{fig:default_behavior_samples}.}
\label{tab:default_behavior_samples}

\end{table}

\clearpage
\section{Low Diversity Originates in Training Data} \label{app:subsec:laion_experiment}

\paragraph{Filtering Captions from LAION.} We aimed to measure the diversity of training images whose captions satisfy two conditions: (1) they mention the concept as an object and not as a modifier (e.g., ``cookie'' but not ``cookie cutter''), and (2) the caption must not mention or imply the attribute of interest (e.g., ``a classic chocolate chip cookie'' implies the cookie is round). We queried LAION using \wimbd{} \citep{wimbd} and sampled 500 captions for each concept.

To efficiently filter the captions, we utilized \gptfouro{} in a few-shot setup. For each caption, we provided the caption text, the concept (e.g., ``cookie''), and the question regarding the attribute of interest (e.g., ``what is the shape of the cookie?''). We instructed \gptfouro{} to analyze each caption and determine whether it satisfies both filtering conditions. The model was prompted to reply with ``yes'' if both conditions are met and ``no'' otherwise.

We then downloaded the images associated with the captions that \gptfouro{} classified as satisfying both conditions. To ensure the reliability of our filtering method, we conducted a human evaluation, achieving an F1 score of 90.3\%. Detailed methodology and results of the human evaluation are provided in \cref{app:human_eval}.

Below is the prompt we use with \gptfouro{} to filter captions from LAION:

\begin{tcolorbox}[myverbatimbox]
\begin{Verbatim}
In this task, you are provided with a caption associated with an image, a concept, and a question. You need to find relevant captions that do not indicate the answer to the question. Your role is two-part. First, determine whether the caption explicitly mentions the concept as a tangible thing, and not an accessory or an item related to the concept. Second, determine if that question can be answered only by reading the caption. If the answer is yes for the first and no for the second, reply with "yes", otherwise reply with "no".

Here are some examples to guide your understanding:

Caption: teapot, glass teapot, Chinese teapot, herbal teapot, teaware
Concept: teapot
Question: What material is the teapot made of (ceramic, metal, glass, etc.)?
Reasoning: The first part is to determine if teapot is mentioned in the prompt. It is the first word in the caption, so it is. The second part is to determine if the question is answerable from the prompt or not. We want to find captions that are not answerable. Since there are mentions of materials in the caption, it is answerable and the answer is no.
Answer: no

Caption: My Sweet Angel Book Store Hyatt Book Store Amazon Books eBay Book Book Store Book Fair Book Exhibition Sell your Book Book Copyright Book Royalty Book ISBN Book Barcode How to Self Book
Concept: book
Question: Is the book dirty or clean?
Reasoning: The caption mentions items related to a book, but not an actual book. The answer is no.
Answer: no

Caption: Perfect reading chair, cozy reading chair, nest chair, my favorite chair, Nest Chair, Cozy Chair, Chair Cushions, Big Chair, Cuddle Chair, Swivel Chair, Relax Chair, Big Comfy Chair, Chaise Chair
Concept: chair
Question: What color is the chair?
Reasoning: The first part is to identify if the caption mentions a chair. It does mention a chair, with various adjectives. The second part is to determine if the question is answerable from the caption. The question asks about the color of the chair, and there is no mention of a chair color. The answer is yes.
Answer: yes

Caption: JIX motorcycle helmet, cross helmet, full helmet, safety helmet
Concept: helmet
Question: Does the helmet have any logos or graphics on it?
Reasoning: The first part is to determine if the caption mentions a helmet. The caption indeed mentions a variety of helmets. The second part is to determine if the question can be answered from the caption alone. There is no information about logos or graphics in the caption, so it is not answerable from the caption alone. The final answer is yes because the answer to the first is yes and the second is no.
Answer: yes

Caption: dust bin, garbage container, recycle bin, trash icon
Concept: bin
Question: What shape is the bin?
Reasoning: The first part is to determine if the caption mentions a bin. The caption mentions a bin, but it also mentions trash icon. This indicates this is not an actual bin, but an icon of a bin. The answer is no.
Answer: no

Caption: Cookie Policy - Cookie Law Compliance [MultiLang..
Concept: cookie
Question: What shape is the cookie?
Reasoning: The first part is to determine if the caption mentions a cookie. The caption mentions cookie policy and cookie law compliance, but not an actual edible cookie, that has a shape. The answer is no.
Answer: no

Caption: Best Cookie Presses - Cookie Press 150PCS Cookie Press Gun with 16 Review
Concept: cookie
Question: Does the cookie have chocolate chips?
Reasoning: The first part is to determine if the caption mentions a cookie or something else. The caption is about cookie press and not actual cookie. The answer is no. 
Answer: no
\end{Verbatim}
\end{tcolorbox}

\section{Human Evaluation} \label{app:human_eval}
\paragraph{Worker selection.} Workers were chosen based on their performance records, requiring them to have a minimum of 5,000 approved HITs and an approval rate above 98\%. They had to achieve a perfect score on a qualification exam before being granted access to the task. An hourly wage of \$15 was provided, ensuring they were fairly compensated for their efforts. In total, 71 unique workers participated in evaluating GRADE and 49 to filter the captions from LAION.

\begin{figure}[h]
    \centering
    \includegraphics[width=\textwidth]{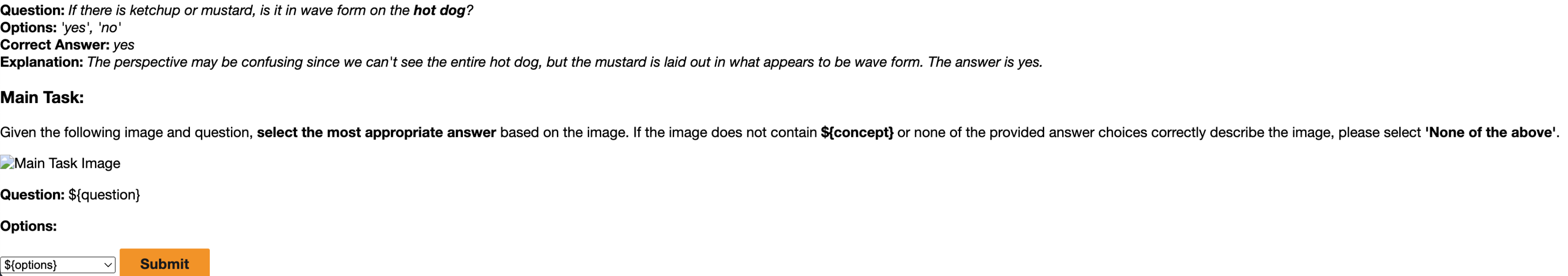}
    \caption{
    A screenshot of the VQA validation task. Workers are provided a question, concept, image, and a set of categories, including ``none of the above'' (options here). Their task is to select the option that answers the question.
    }
    \label{fig:amt_vqa_ui}
\end{figure}

\begin{figure}[h]
    \centering
    \includegraphics[width=\textwidth]{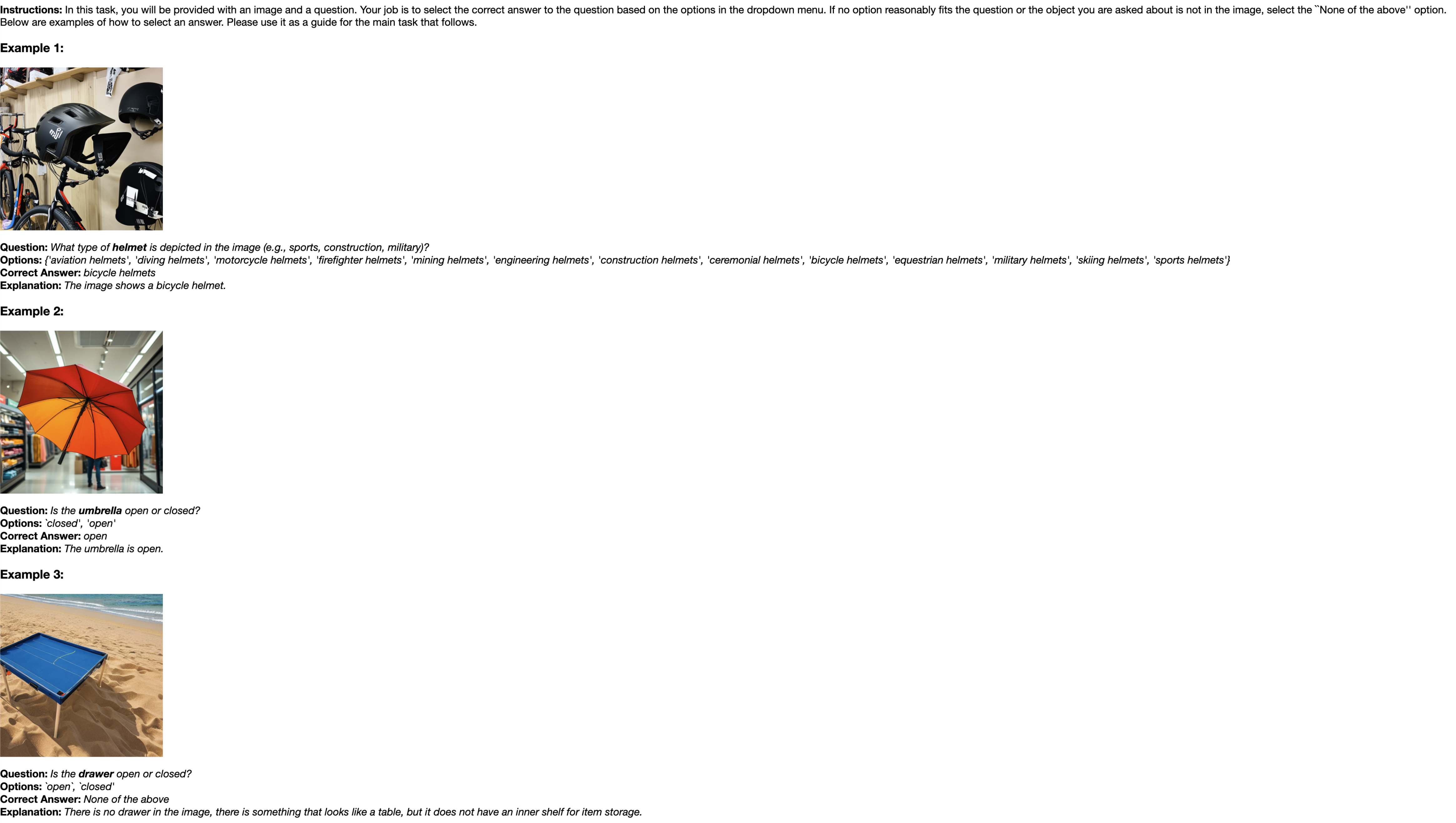}
    \caption{
    3 out of 10 examples provided to workers as aid to complete their visual question answering task. 
    }
    \label{fig:amt_vqa_examples}
\end{figure}

\begin{figure}[h]
    \centering
    \includegraphics[width=\textwidth]{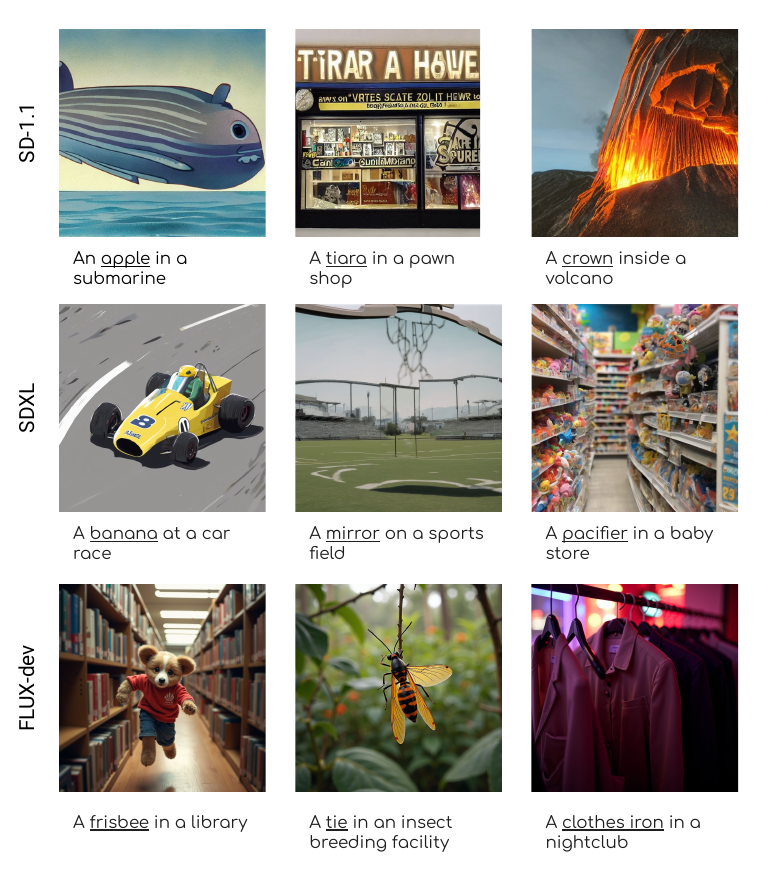}
    \caption{
    A sample of images marked with ``none of the above'', as a result of not including the concept (underlined) in the image.
    }
    \label{fig:none_of_the_above_t2i}
\end{figure}

\begin{figure}[h]
    \centering
    \includegraphics[width=\textwidth]{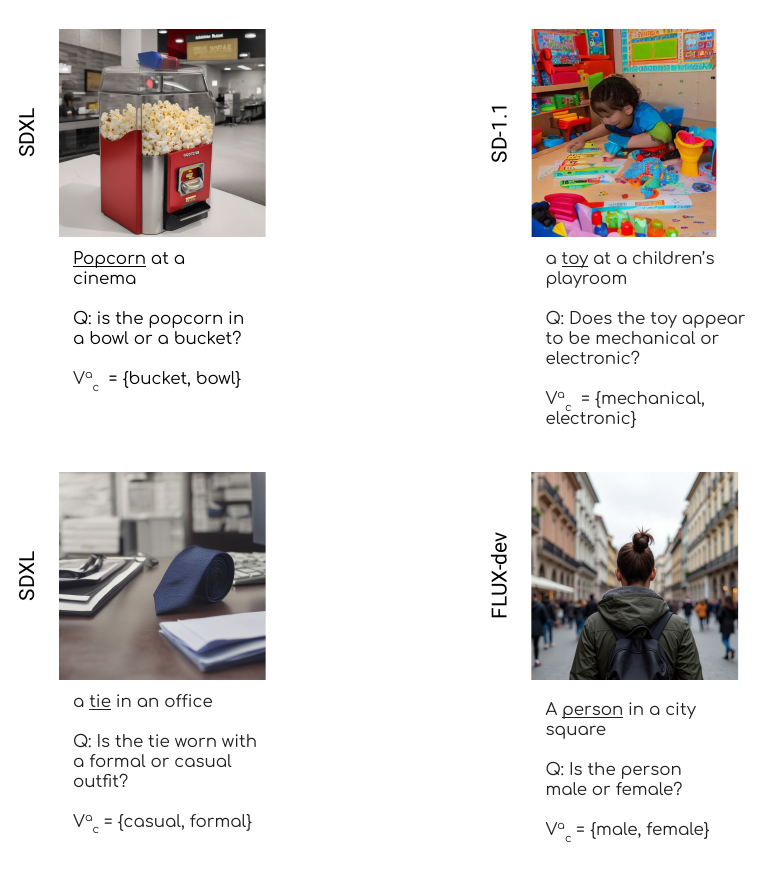}
    \caption{
    A sample of images marked with ``none of the above''. The top row exhibits cases where the attribute value is not in  \( \mathcal{V}_c^a \). The bottom row exhibits cases where the question cannot be answered just from viewing the image. The concept in each prompt is underlined. 
    }
    \label{fig:none_of_the_above_category_and_vqa}
\end{figure}

\paragraph{Validating \grade{}.} To validate the VQA \cref{subsec:validating_grade}, we run an AMT crowdsourcing task where the worker is provided with a question, concept, image, and attribute values, and is requested to select the attribute value that best matches
the question and image. The UI for this task can be viewed in \cref{fig:amt_vqa_ui} with examples in \cref{fig:amt_vqa_examples}. A sample of cases from our attribute values coverage validation (validation of step (b)) is available in \cref{fig:none_of_the_above_t2i} and \cref{fig:none_of_the_above_category_and_vqa}.

\begin{figure}[h]
    \centering
    \includegraphics[width=\textwidth]{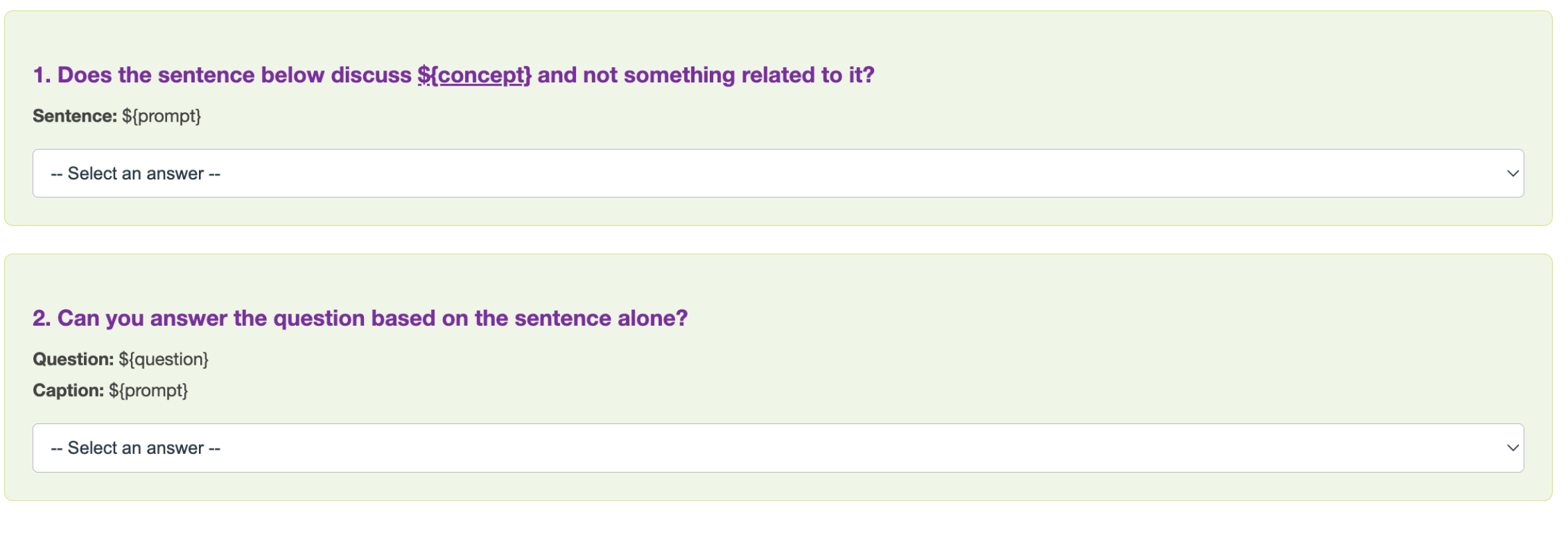}
    \caption{
    A screenshot of the caption filtering validation task. Workers are provided a caption, two questions, and a concept. Their task is to read the caption and answer the questions.}
    \label{fig:caption_filtering_ui}
\end{figure}

\begin{figure}[h]
    \centering
    \includegraphics[width=\textwidth]{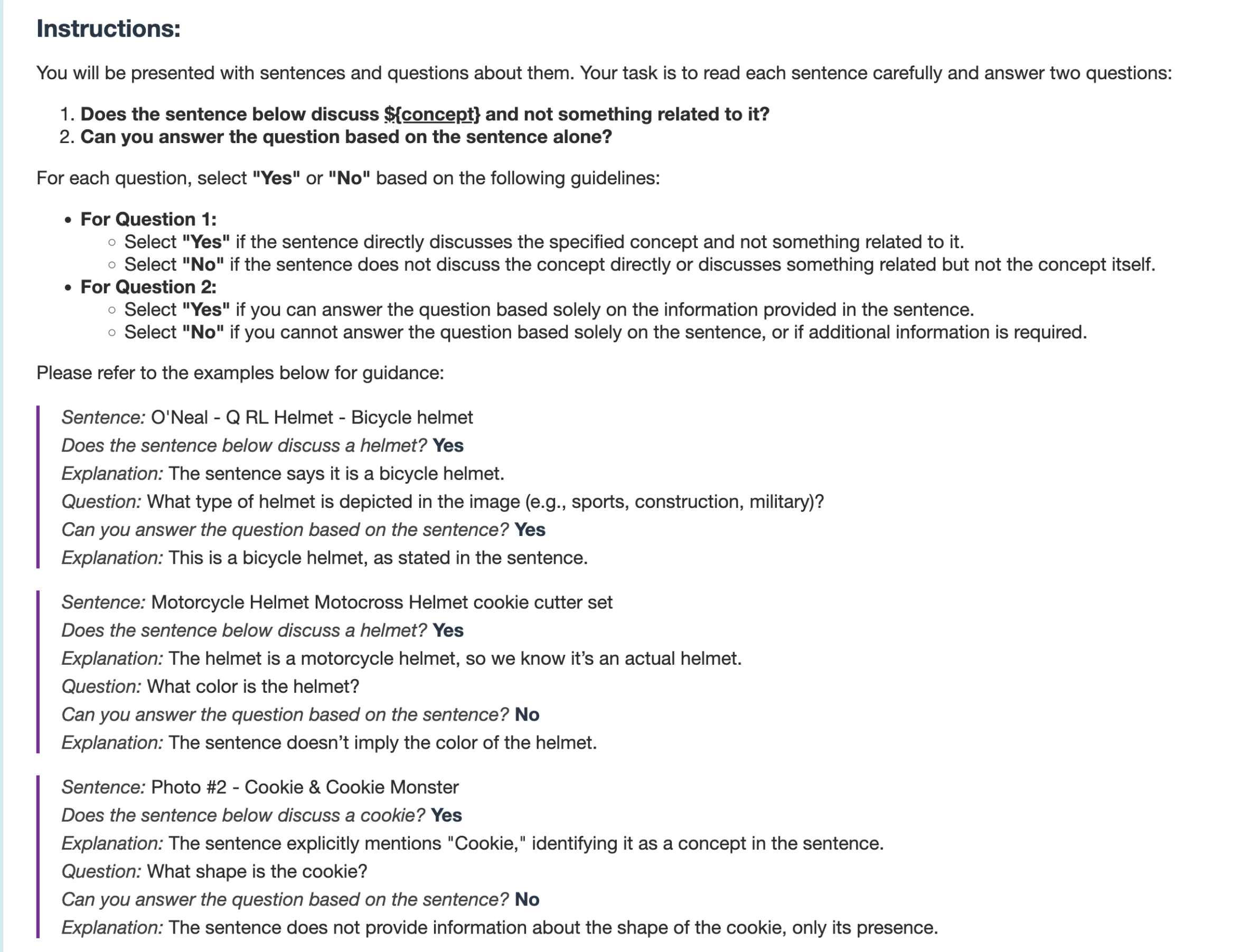}
    \caption{
    3 out of 10 examples provided to workers as aid to complete their caption filtering task. 
    }
    \label{fig:caption_filtering_examples}
\end{figure}

\paragraph{Validating filtering of captions from LAION.} To assess the effectiveness of our \gptfouro{}-based caption filtering method described in \cref{sec:training_data_exp}, we conducted an Amazon Mechanical Turk (AMT) crowdsourcing task. We sampled 1,000 captions from LAION, ensuring an equal distribution of 500 captions that met the filtering criteria and 500 that did not. Workers were instructed to evaluate whether each caption (1) explicitly mentioned the concept as the main object rather than as a modifier (e.g., ``cookie'' instead of ``cookie cutter'') and (2) the caption must not mention or imply the attribute of interest (e.g., ``a classic chocolate chip cookie'' implies the cookie is round). Each example was reviewed by three independent workers, and the majority decision was taken as the final label. Our automated filtering method achieved a recall of 85.8\% and a precision of 95.4\%, resulting in an F1 score of 90.3\%, which indicates a high level of agreement with human judgments. These findings demonstrate that \gptfouro{} is a reliable tool for automated caption filtering. Additional details about the user interface and example cases are provided in \cref{fig:caption_filtering_ui} and \cref{fig:caption_filtering_examples}, respectively.

\clearpage

\section{Prompts in \grade{}} \label{app:prompts_in_grade}

\subsection{Concept Collection} \label{app:concept_collection}
To collect a list of diverse concepts, we prompt \gptfouro{} \citep{gpt4} with the following: 

\begin{tcolorbox}[myverbatimbox]
\begin{Verbatim}
Provide a CSV of 100 unique concepts, like the example below.
concept_id is an enumeration that begins from 0. 
Choose concepts that are easy to visually verify for a VQA model.

concept_id,concept
0, an ice cream
1, a cake
2, a suitcase
3, a clock
\end{Verbatim}
\end{tcolorbox}

\subsection{Prompt generation}
The following prompt was used to generate common prompts:

\begin{tcolorbox}[myverbatimbox]
\begin{Verbatim}
Please suggest three typical settings for the concept below.  
Note that the output should be a list of strings.

Here's an example:
Concept: a cake
Prompts: [
"a cake in a bakery, 
"a cake at a birthday party", 
"a cake at a swimming pool"
]

Concept: {concept}
\end{Verbatim}
\end{tcolorbox}

This one was used to generate uncommon prompts:

\begin{tcolorbox}[myverbatimbox]
\begin{Verbatim}
Please suggest three atypical settings for the concept below.  
Note that the output should be a list of strings.

Here's an example:
Concept: a cake
Prompts: [
"a cake on a weight loss clinic, 
"a cake at a gym", 
"a cake at a swimming pool"
]

Concept: {concept}
\end{Verbatim}
\end{tcolorbox}

\subsection{Attribute Generation} 
\grade{} first analyzes the specific attributes of the concept provided in the prompt, and then generates questions that can be used to count the occurrences of attribute values in images. Below is the prompt we used with \gptfouro{}.

\begin{tcolorbox}[myverbatimbox]
\begin{Verbatim}
Help me ask questions about images that depict certain concepts.

I will provide you a concept.
Your job is to analyze the concept's typical attributes 
and ask simple questions that can be answered by viewing the image. 

Here's an example:

concept: 
a cake
attributes: 
cakes can be made in different flavors, shapes, 
and can have multiple tiers.

questions: 
1. Is the cake eaten?
2. Does the cake have multiple tiers?
3. In what flavor is the cake?
4. What is the shape of the cake?
5. Does the cake show any signs of fruit on the outside or 
suggest a fruit flavor?

Now that you understand, let's begin. 

concept: {c}
\end{Verbatim}
\end{tcolorbox}

\subsection{Attribute Values Generation}
To generate attribute values \( \tilde{\mathcal{V}}_c^a\) for \( \tilde{P}_{V \mid a,c}\), we provide \gptfouro{} \citep{gpt4} with a concept, a question, and a prompt. \gptfouro{} then outputs a list of attribute values that can match the question (attribute). The process is performed for all prompts mentioning the concept. The sets are then unified with similar answers removed (e.g., ``motorbike helmets'' is removed, because ``motorcycle helmets'' already exists). The result of the unification is \( \tilde{\mathcal{V}}_c^a\).

\begin{tcolorbox}[myverbatimbox]
\begin{Verbatim}
I have a question that is asked about an image.  I will provide you with the question and a caption of the image. 
Your job is to first analyze the description of the image and the question, then, hypothesize plausible answers that can surface from viewing the image. Do not write anything other than the answer.
Then, I need you to list the plausible answers in a list, just like in the example below. For example,
Caption: a helmet in a bike shop
Question: What type of helmet is depicted in the image? 
Plausible answers: ["motorcycle helmets",
            "bicycle helmets",
            "football helmets",
            "construction helmets",
            "military helmets",
            "firefighter helmets",
            "rock climbing helmets",
            "hockey helmets"]
Now your turn. 
Caption: {caption}
Question: {question} 
Plausible answers:


I have a question that is asked about an image. I will provide you with the question and a caption of the image. Your job is to first carefully read the question and analyze, then hypothesize plausible answers to the question assuming you could examine the image (instead, you examine the caption). The answers should be in a list, as in the example below. Do not write anything other than the plausible answers. 

Example:
Caption: a helmet in a bike shop
Question: What type of helmet is depicted in the image? 
Plausible answers: ["motorcycle helmets",
            "bicycle helmets",
            "football helmets",
            "construction helmets",
            "military helmets",
            "firefighter helmets",
            "rock climbing helmets",
            "hockey helmets"]
Now your turn. 
Caption: {caption}
Question: {question} 
Plausible answers:
\end{Verbatim}
\end{tcolorbox}

\subsection{Generating answers}
We use \gptfouro{} to answer the generated questions with $1,000$ as max tokens and temperature 0. We use the Structured Outputs feature \citep{openai_structured_outputs_api} to map the natural language answers to attribute values in a single step. Our prompt is straightforward:

\begin{tcolorbox}[myverbatimbox]
\begin{Verbatim}
Answer the following question with one of the categories. To come up with the correct answer, carefully analyze the image and think step-by-step before providing the final answer.

Question: {question}
Categories:{categories}
Selection:
\end{Verbatim}
\end{tcolorbox}

\end{document}